\pdfoutput=1

\documentclass[11pt]{article}

\usepackage[preprint]{acl}

\usepackage{times}
\usepackage{latexsym}

\usepackage[T1]{fontenc}

\usepackage[utf8]{inputenc}

\usepackage{microtype}

\usepackage{inconsolata}

\usepackage{listings}
\usepackage{xcolor}
\usepackage{graphicx}
\usepackage{caption}
\usepackage{subcaption}
\usepackage{arydshln}
\usepackage{enumitem}
\usepackage{multirow}
\usepackage{xspace}
\usepackage{float}

\newcommand{\ie}{\textit{i}.\textit{e}.}
\newcommand{\eg}{\textit{e}.\textit{g}.}
\newcommand{\etc}{\textit{etc}.\xspace}

\newcommand{\bench}{\texttt{CloudAPIBench}\xspace}
\definecolor{verylightgray}{gray}{0.85}
\definecolor{lightred}{rgb}{1, 0.8, 0.8}
\newcolumntype{a}{>{\columncolor{verylightgray}}c}

\definecolor{codegreen}{rgb}{0,0.6,0}
\definecolor{codegray}{rgb}{0.5,0.5,0.5}
\definecolor{codepurple}{rgb}{0.58,0,0.82}
\definecolor{backcolour}{rgb}{0.95,0.95,0.92}
\lstdefinestyle{systemprompt}{
    backgroundcolor=\color{backcolour},   
    commentstyle=\color{codegreen},
    keywordstyle=\color{magenta},
    numberstyle=\tiny\color{codegray},
    stringstyle=\color{codepurple},
    basicstyle=\ttfamily\footnotesize,
    breakatwhitespace=false,         
    breaklines=true,                 
    captionpos=b,                    
    keepspaces=true,                 
    numbers=left,                    
    numbersep=5pt,                  
    showspaces=false,                
    showstringspaces=false,
    showtabs=false,                  
    tabsize=2
}

\title{On Mitigating Code LLM Hallucinations with API Documentation}

\author{Nihal Jain \\ \texttt{nihjain@amazon.com} \\  Amazon Web Services, USA \And
        Robert Kwiatkowski\thanks{Work done while author was at Amazon.} \\ \texttt{robert.kwiatkowski@columbia.edu} \AND
        Baishakhi Ray \\ \texttt{rabaisha@amazon.com} \\  Amazon Web Services, USA \And
        Murali Krishna Ramanathan \\ \texttt{mkraman@amazon.com} \\  Amazon Web Services, USA \And
         Varun Kumar \\ \texttt{kuvrun@amazon.com} \\  Amazon Web Services, USA}

\begin{document}
\maketitle
\begin{abstract}
In this study, we address the issue of API hallucinations in various software engineering contexts. We introduce \bench, a new benchmark designed to measure API hallucination occurrences. \bench also provides annotations for frequencies of API occurrences in the public domain, allowing us to study API hallucinations at various frequency levels. Our findings reveal that Code LLMs struggle with low frequency APIs: for \eg, GPT-4o achieves only $38.58$\% valid low frequency API invocations. We demonstrate that Documentation Augmented Generation (DAG) significantly improves performance for low frequency APIs (increase to $47.94$\% with DAG) but negatively impacts high frequency APIs when using sub-optimal retrievers (a $39.02$\% absolute drop). To mitigate this, we propose to intelligently trigger DAG where we check against an API index or leverage Code LLMs' confidence scores to retrieve only when needed. We demonstrate that our proposed methods enhance the balance between low and high frequency API performance, resulting in more reliable API invocations ($8.20$\% absolute improvement on \bench for GPT-4o).
\end{abstract}

\section{Introduction}

\begin{figure*}
    \centering
    \begin{subfigure}{0.44\textwidth}
        \includegraphics[width=\textwidth]{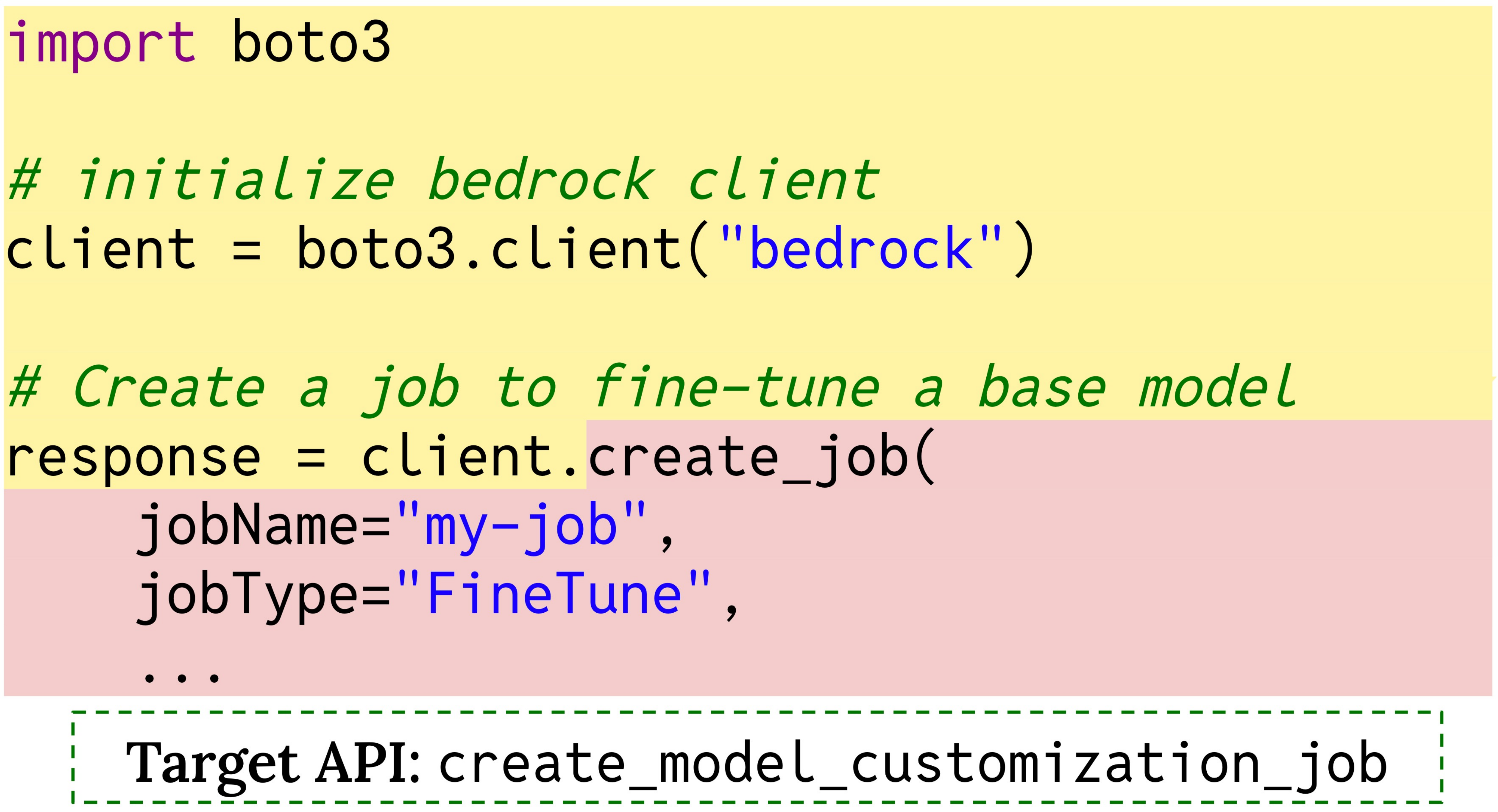}
    \end{subfigure}
    \hfill
    \begin{subfigure}{0.53\textwidth}
        \includegraphics[width=\textwidth]{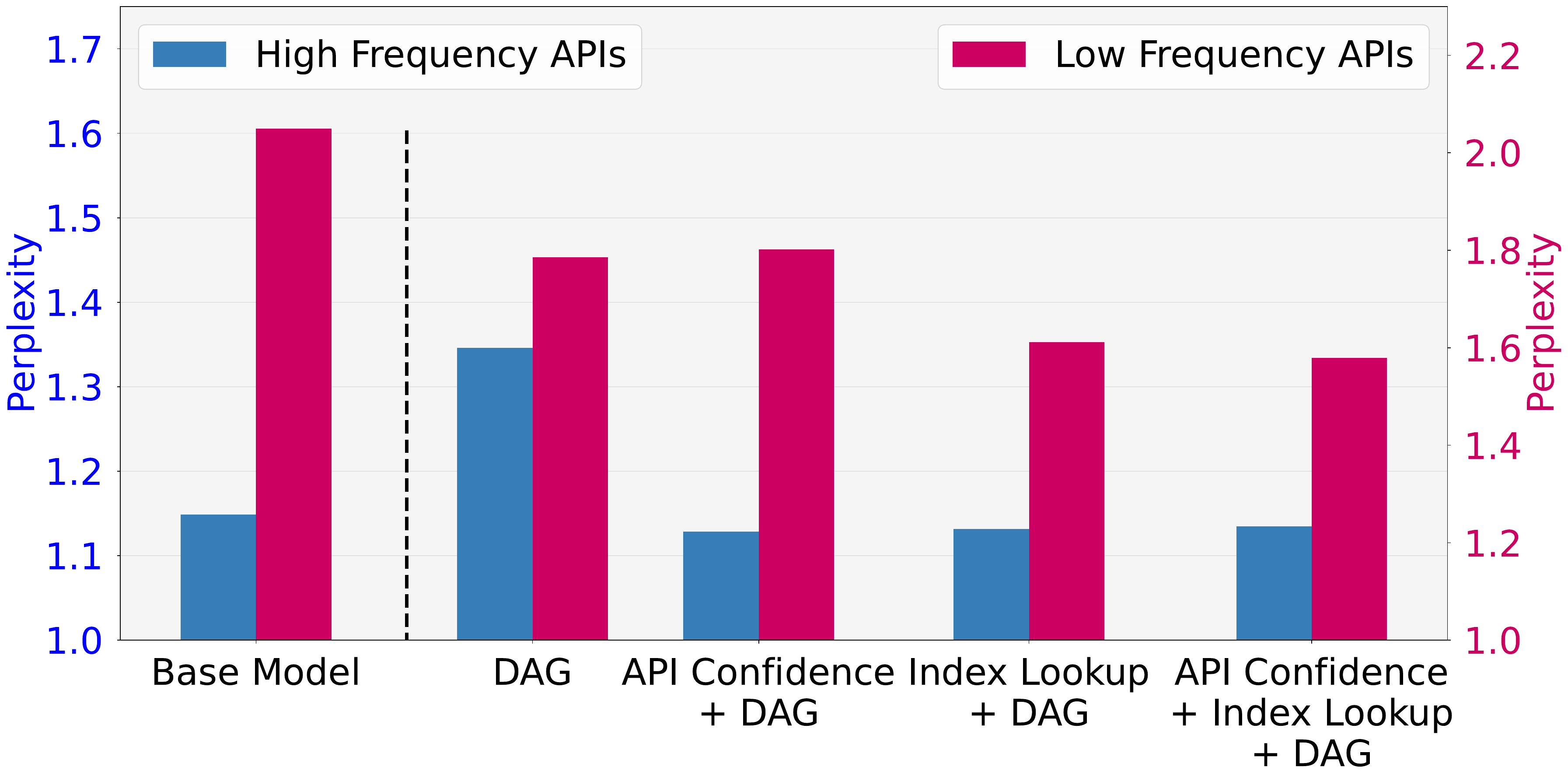}
    \end{subfigure}
    \caption{\textbf{Introduction.} \textbf{(Left)}~A \bench task (yellow) and StarCoder2-15B's response (red) are displayed. The target is a recently released AWS API~\citep{bedrock2023}, \ie, a \textit{low frequency} API. Due to limited training on such APIs, the Code LLM hallucinates a non-existent API invocation. \textbf{(Right)}~Given a prompt from \bench, we measure the perplexity of the target API tokens using StarCoder2-15B (\textit{lower is better}). The base model handles high frequency APIs well but falters with low frequency ones. While DAG (with imperfect retrievers) improves low frequency API performance, it hurts high frequency API performance due to irrelevant augmentations. This paper's methods and analyses address this limitation of DAG.}
    \label{fig:introduction}
\end{figure*}

Programmers frequently utilize third-party Application Programming Interfaces (APIs) as foundational elements for new software development, particularly in domains like cloud services, web and mobile development, e-commerce, FinTech, and data analytics. These APIs offer essential functionalities, enabling developers to create robust and feature-rich applications efficiently.

Large Language Models for code generation (Code LLMs) are being increasingly used by programmers~\citep{peng2023impact,chen2021evaluating,dakhel2023github}. However, these models can generate incorrect API-related code, known as \textit{API hallucinations}~\citep{liu2024exploring}, especially when under-trained on certain under-represented APIs -- referred to as \textit{low frequency} APIs (see Figure~\ref{fig:introduction}~(left)). This problem is exacerbated by the constant evolution of APIs, including frequent updates and deprecation of existing APIs~\cite{mcdonnell2013empirical}. Consequently, new, updated, or infrequently used APIs are more prone to hallucinations. To systematically measure the prevalence of such hallucinations, we introduce \textbf{\bench}, a benchmark specifically designed to evaluate API hallucinations, focusing on APIs from major cloud service providers like Amazon Web Services (AWS) and Microsoft Azure.

Next, we explore mitigation strategies for API hallucinations.
When uncertain about API usage, human developers frequently rely on API documentation. Likewise, we hypothesize that Code LLMs should consult these resources under uncertainty.
Hence, to address API hallucinations, we adopt retrieval augmented generation with documentation, \ie, \textbf{Documentation Augmented Generation (DAG)}, which has shown early promise~\citep{zhou2023docprompting, patil2023gorilla}. 

However, DAG may be unnecessary when APIs are stable or well-covered in the model's training data (\ie, \textit{high frequency} APIs)---we find that DAG with suboptimal retrievers indeed degrades performance for high frequency APIs, supporting the observation that LLMs are sensitive to irrelevant information~\citep{shi2023large, yoran2023making}. As such, we also present two simple yet effective strategies that can be easily adapted with DAG to address such pitfalls.

Figure~\ref{fig:introduction}~(right) demonstrates how the frequency of an API's occurrence in the public domain affects Code LLMs. We analyze the perplexity of StarCoder2-15B (base model)~\citep{lozhkov2024starcoder} on API tokens across two frequency groups: \textit{low} ($\leq 10$ occurrences in training data: The Stack v2) and \textit{high} ($\geq 100$ occurrences), with detailed frequency descriptions in Section~\ref{sec:cloudapibench}. The base model excels with high frequency APIs but struggles with low frequency ones. While DAG enhances performance for low frequency APIs, it compromises high frequency API performance due to occasional irrelevant augmentations from suboptimal retrievers: these distract the Code LLM's reliance on its internal knowledge, which is sufficient for high frequency APIs. To address DAG's limitations, we explore methods such as inspecting model confidence scores~\citep{jiang2023active, varshney2023stitch} and validating model generations against an API index before retrieval. These strategies effectively mitigate DAG's drawbacks, enhancing the reliability of Code LLMs.

We outline our key contributions and the structure of the paper as follows:

\begin{itemize}[leftmargin=*]
    \itemsep0em
    \item We introduce \bench to systematically study real-world API hallucinations; this benchmark evaluates API hallucinations across major cloud SDK APIs, \ie, those by AWS and Azure (\textbf{Section~\ref{sec:api-hallucinations-cloudapibench}}).
    \item We present a thorough study of DAG to enhance \bench performance, identifying the parts of documentation that reduce hallucinations and quantifying the impact of other retrieval components on model efficacy (\textbf{Section~\ref{sec:documentation-augmented-generation}}).
    \item We characterize scenarios where DAG may degrade performance and discuss selective retrieval methods to improve Code LLMs' utilization of documentation (\textbf{Section~\ref{sec:improving-dag}}).
\end{itemize}

We believe this is the first work to measure and characterize real-world API hallucinations for various Code LLMs and explore strategies to mitigate this issue through documentation.
\section{API Hallucinations \& \bench} \label{sec:api-hallucinations-cloudapibench}

We first comment on the impact of API hallucinations in Section~\ref{sec:api-hallucinations} and introduce \bench to measure these in Section~\ref{sec:cloudapibench}.

\subsection{Impact of API Hallucinations} \label{sec:api-hallucinations}

\citet{liu2024exploring} identify that API hallucinations constitute up to $15\%$ of all hallucinations in state-of-the-art Code LLMs, influencing cloud software engineering where code is API-intensive. These hallucinations can propagate errors,  creating a snowball effect~\citep{zhang2023language}. For \eg, a hallucinated API call can lead to hallucinated handling of its response in subsequent code segments, compounding the problem~\citep{ding-etal-2023-static}. Such incorrect API usage can also introduce security vulnerabilities, like improper data handling, which may lead to attacks or data breaches~\citep{pearce2021asleep}. As adoption of Code LLMs grows, the cognitive burden on developers increases~\citep{barke2022grounded}, as they must trace and rectify both the initial hallucination and all affected code segments. Given this severe impact of API hallucinations, it is critical to explore effective methods for their detection and mitigation, as we study in this work.

\subsection{\bench} \label{sec:cloudapibench}

\begin{figure}
    \centering
    \includegraphics[width=\columnwidth]{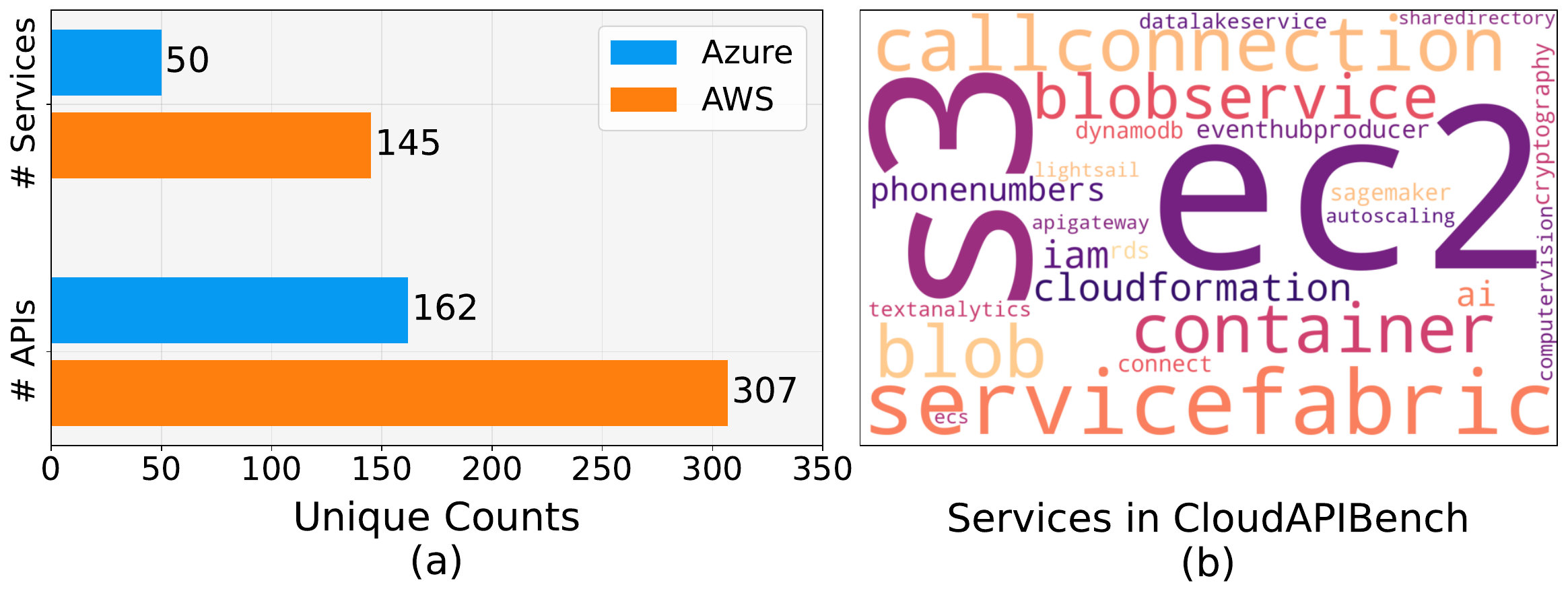}
    \caption{\textbf{Composition of \bench.} (a)~The benchmark comprises diverse APIs from various AWS and Azure services. (b)~Word cloud visualizing the services in \bench; from AWS \textbf{\texttt{s3}} to Azure \textbf{\texttt{computervision}}, \bench comprises many cloud-based software engineering use-cases.}
    \label{fig:cloudapibench-dist}
\end{figure}

Current benchmarks evaluate various programming skills such as problem solving~\citep{chen2021evaluating, austin2021program}, repository-level coding~\citep{ding2023crosscodeeval, liu2023repobench}, and tool usage~\citep{patil2023gorilla, basu2024apiblend}. However, a comprehensive benchmark for assessing real-world API hallucinations in software engineering remains absent. To address this, we introduce \bench, a benchmark designed to evaluate Code LLMs' abilities to invoke cloud APIs.

\noindent \textbf{Composition.} \bench is a Python benchmark comprising $622$ \textit{synthetic tasks} to prevent data leakage with Code LLMs. Each task requires invoking a specific cloud API from providers like AWS and Azure, reflecting practical software engineering scenarios. Task prompts include imports, variable declarations, and a developer-style comment describing the API's purpose, stopping just before the API invocation. Figure~\ref{fig:introduction}~(left) illustrates a sample task and a model response (demonstrating an API hallucination). Figure~\ref{fig:cloudapibench-dist} presents a detailed task distribution, showing that \bench captures diverse API invocation scenarios to evaluate Code LLMs comprehensively.

\noindent \textbf{API Frequency Annotations.} \bench also contains the \textit{API frequency} for APIs, \ie, how often they occur in The Stack v2~\citep{lozhkov2024starcoder}. As The Stack v2 is one of the largest open code pre-training datasets, we assume that our API frequencies approximates the distribution of APIs in public sources. Hence, this can be used to explore the relationship between hallucination rates and API frequencies for various Code LLMs.

To enhance interpretability, we classify API frequencies into three categories: \textit{Low} ($0-10$ occurrences), \textit{Medium} ($11-100$), and \textit{High} ($\ge 101$). Since this treats APIs within the same class as identical, we minimize confounding factors (such as invocation complexity) by selecting diverse APIs within each class. This approach parallels the categorization of concepts based on popularity or pre-training frequencies~\citep{razeghi2022impact, mallen2022trust}. To our knowledge, this is the first granular analysis of a Code LLM's pre-training corpus. Detailed API frequency distributions in \bench are provided in Appendix~\ref{sec:appendix-cloudapibench-composition}.

\noindent \textbf{Construction.} We construct \bench with the goal of scaling coverage to multiple APIs from various providers. First, we source API specifications from official documentation to index their correct usage. Next, we determine each API's frequency in The Stack v2 by counting function definitions and calls with the same names as the APIs in relevant files. We select APIs for \bench while ensuring diversity of invocation complexity and frequency. Using Claude 3 Sonnet~\citep{Claude3}, we convert API descriptions into developer-style comments, and create prompts with necessary imports, declarations, and a descriptive comment before the API call. We provide elaborate details of this process in Appendix~\ref{sec:appendix-cloudapibench-construction}, and present more samples from \bench in Appendix~\ref{sec:appendix-cloudapibench-samples}.

\begin{figure}
    \centering
    \includegraphics[width=\columnwidth]{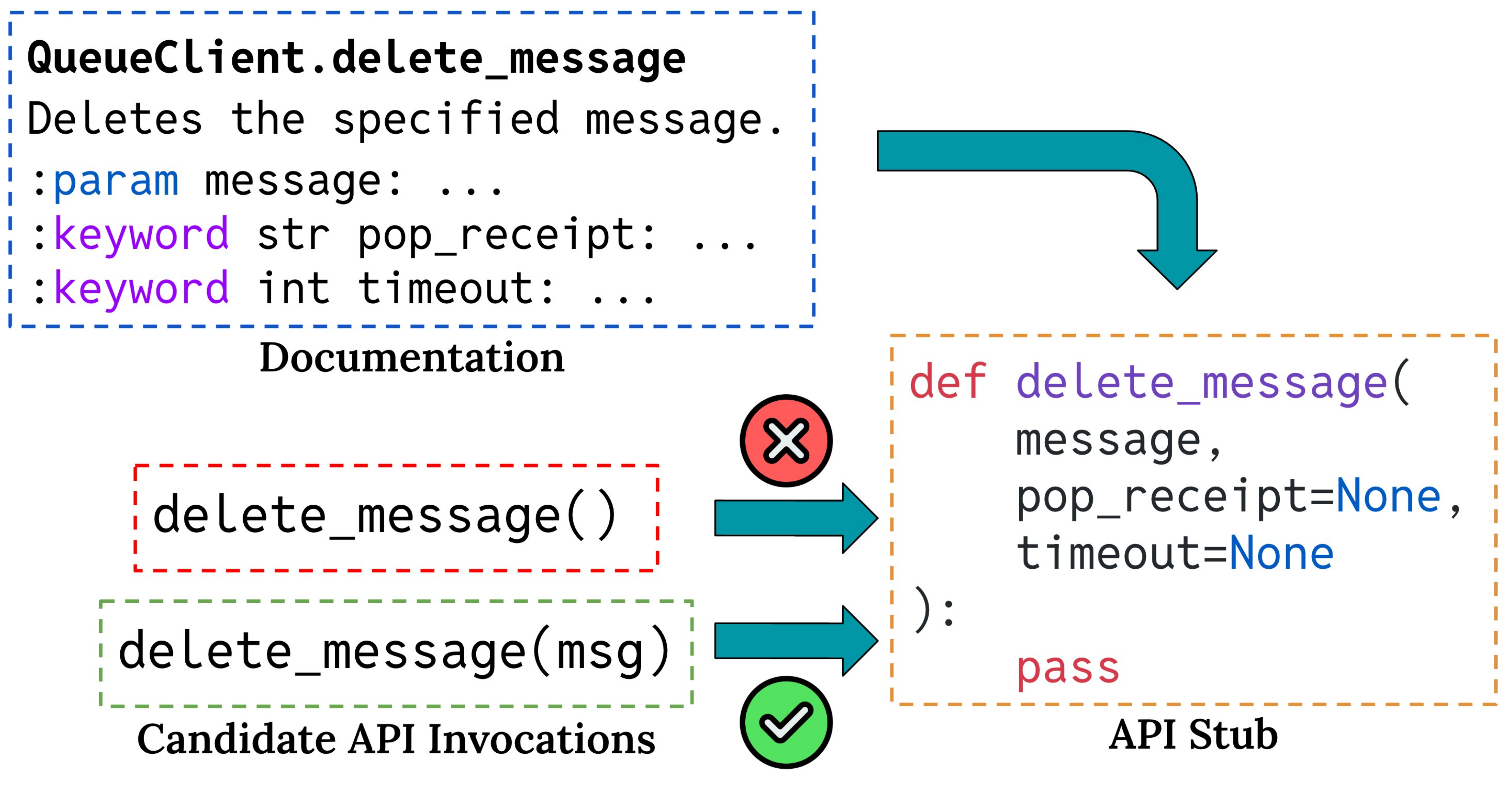}
    \caption{\textbf{Valid API Invocation.} Using the API documentation, we create an API stub to capture correct usage. A candidate invocation is valid if it successfully binds to the stub. Here, \texttt{delete\_message} requires \textit{at least} one required argument for successful binding.}
    \label{fig:valid-api-invocation}
\end{figure}

\noindent \textbf{Evaluation Metric.} We introduce the \textbf{valid API invocation} metric, which verifies if an API is invoked according to its syntax. We obtain this syntax by tracking the API's arguments and whether they are required. The metric is computed as follows: we create a dummy function mirroring the API's signature (\ie, API stub~\cite{zhu2023stubcoder}). A candidate invocation is tested against this stub, and only successful bindings indicate validity. This evaluation method bypasses the intricacies of static analysis~\citep{patil2023gorilla} and is more robust than string matching~\citep{ding2023crosscodeeval}, ensuring reliable and scalable evaluations. Figure~\ref{fig:valid-api-invocation} illustrates this process. See Appendix~\ref{sec:appendix-cloudapibench-evaluation} for more details. 

\begin{table*}
\centering
\resizebox{0.8\textwidth}{!}{%
\renewcommand{\arraystretch}{1.2}
    \begin{tabular}{c|c|cccccc|}
    \hline
    \hline
    \multirow{2}{*}{\textbf{Model}} & \textbf{HumanEval} & \multicolumn{6}{c}{\textbf{\bench}} \\
    & \textit{pass@1} & \textit{High Frequency} & \textit{Medium Frequency} & \textit{Low Frequency} \\
    \hline
    \hline
    StarCoder2-3B                                      & $31.44$ & $84.39$ & $37.33$ & \cellcolor[RGB]{255, 230, 230}$11.61$ \\
    StarCoder2-7B                                      & $34.09$ & $86.34$ & $47.33$ & \cellcolor[RGB]{255, 230, 230}$9.36$ \\
    StarCoder2-15B                                     & $44.15$ & $88.78$ & $57.33$ & \cellcolor[RGB]{255, 230, 230}$24.72$ \\
    \hdashline
    Google CodeGemma-2B                                & $27.28$ & $79.51$ & $26.67$ & \cellcolor[RGB]{255, 230, 230}$4.49$ \\
    Google CodeGemma-7B                                & $40.13$ & $87.80$ & $52.67$ & \cellcolor[RGB]{255, 230, 230}$12.36$ \\
    \hdashline
    IBM Granite-Code-3B                                & $--$ & $83.41$ & $44.67$ & \cellcolor[RGB]{255, 230, 230}$17.23$ \\
    IBM Granite-Code-8B                                & $--$ & $85.85$ & $62.67$ & \cellcolor[RGB]{255, 230, 230}$28.09$ \\
    IBM Granite-Code-20B                               & $--$ & $87.80$ & $69.33$ & \cellcolor[RGB]{255, 230, 230}$32.21$ \\
    \hdashline
    DeepSeekCoder-1.3B                                 & $32.13$ & $79.02$ & $22.67$ & \cellcolor[RGB]{255, 230, 230}$5.24$ \\
    DeepSeekCoder-6.7B                                 & $45.83$ & $88.78$ & $52.00$ & \cellcolor[RGB]{255, 230, 230}$13.48$ \\
    DeepSeekCoder-33B                                  & $52.45$ & $90.24$ & $70.00$ & \cellcolor[RGB]{255, 230, 230}$34.83$ \\
    \hdashline
    GPT-4o                                             & $90.20$ & $93.66$ & $78.67$ & \cellcolor[RGB]{255, 230, 230}$38.58$ \\
    \hline
    \hline
    
    \end{tabular}
}%
\caption{\textbf{Results on \bench.} We present Valid API Invocation (\%) results on \bench for various Code LLMs, categorized by API frequency in The Stack v2. For comparison, we also include HumanEval~\citep{chen2021evaluating} results from \citet{bigcode-leaderboard} and \citet{gpt4o}. While Code LLMs excel on high-frequency APIs, their performance drops severely on low-frequency APIs, despite strong results on general programming tasks like HumanEval.}
\label{tab:no-augmentation-results}
\end{table*}

\subsection{Evaluation \& Results}

\noindent \textbf{Models.} We evaluate the following recent Code LLMs (and sizes) on \bench: StarCoder2-\{3B, 7B, 15B\}~\citep{lozhkov2024starcoder}, DeepSeekCoder-\{1.3B, 6.7B, 33B\}~\citep{guo2024deepseekcoder}, Google CodeGemma-\{2B, 7B\}~\citep{codegemma_2024}, IBM Granite-Code-\{3B, 8B, 20B\}~\citep{mishra2024granite} and GPT-4o~(\texttt{gpt-4o-2024-05-13})~\citep{gpt4o}.

\noindent \textbf{Inference.} We use greedy decoding, generating one sequence per task up to $256$ tokens, and post-process until the first function call; this is evaluated for validity as detailed in Section~\ref{sec:cloudapibench}. This strategy is used throughout this work consistently. For instruction-tuned models, we specify a system prompt indicating the model to generate only the API invocation (see Appendix~\ref{sec:appendix-cloudapibench-evaluation}).

\noindent \textbf{Results.} Table~\ref{tab:no-augmentation-results} presents the performance of all models on \bench and HumanEval (for a reference of generic performance). Key observations include:

\noindent -- \textit{API Hallucinations.} All Code LLMs exhibit API hallucinations to a certain degree. These primarily occur due to (1) usage of non-existing APIs, (2) incorrect usage of the target API or, (3) usage of incorrect existing APIs. We illustrate these failure cases in Appendix~\ref{sec:appendix-cloudapibench-results}.

\noindent -- \textit{API Frequency Trends.} A strong correlation exists between API frequency and valid API invocations: high frequency APIs yield fewer hallucinations, while low frequency APIs result in more. While this is expected, this trend verifies the applicability of our API frequency annotations.

\noindent -- \textit{Low Frequency APIs.} Despite strong performance on high frequency APIs and generic benchmarks, all models exhibit high hallucination rates for low frequency APIs. This disparity highlights the value of \bench in pinpointing scenarios where Code LLMs are prone to hallucinate. We dive deeper into various low frequency API failure cases for all models in Appendix~\ref{sec:appendix-cloudapibench-low-freq-deep-dive}. \\

\noindent Given the poor performance on low frequency APIs, we now explore the use of documentation to enhance performance on \bench.

\section{Documentation Augmented Generation (DAG)} \label{sec:documentation-augmented-generation}

In this section, we see how DAG enhances performance on \bench. We first outline the key components of DAG: augmentation design, retrieval index and retriever, in Section~\ref{sec:dag-setup}. Subsequently, we discuss how different design choices affect downstream performance in Section~\ref{sec:dag-experiments}.

\subsection{Setup} \label{sec:dag-setup}

\begin{figure}
    \centering
    \includegraphics[width=1.05\columnwidth]{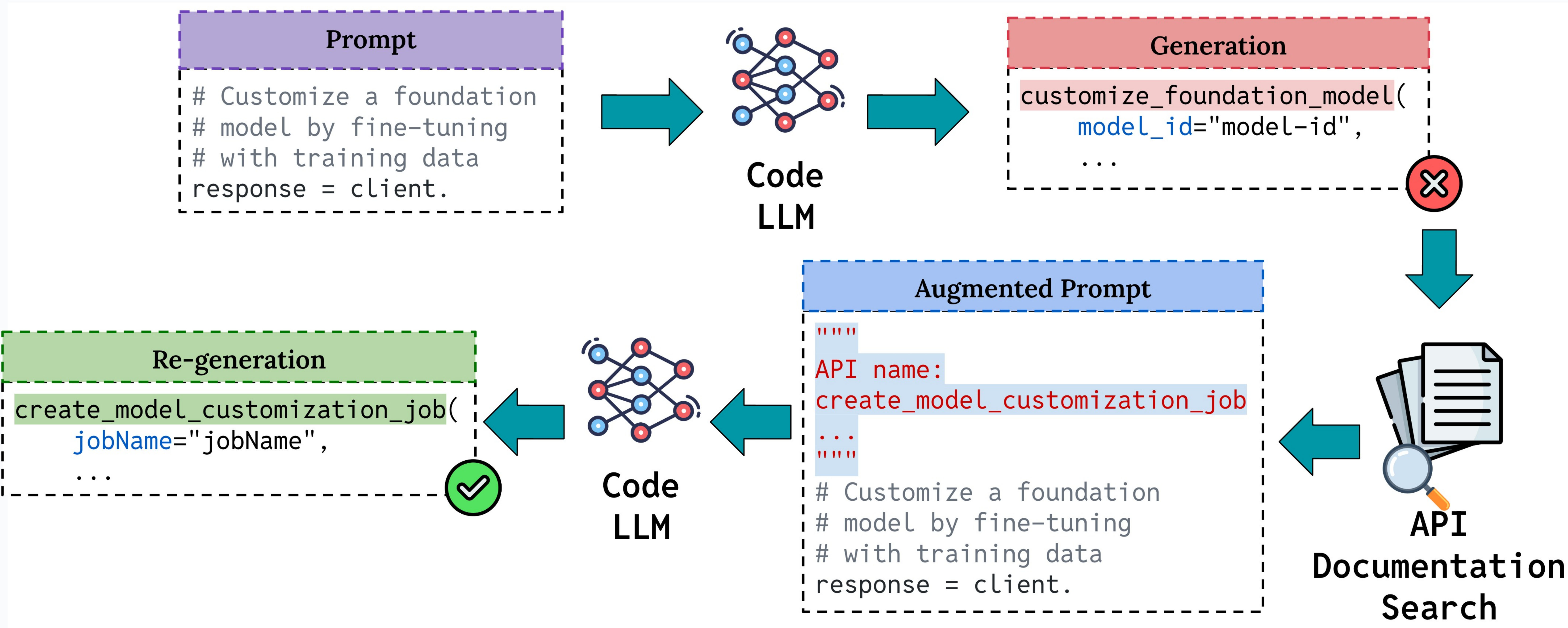}
    \caption{\textbf{DAG Overview.} Starting with a \bench task, we sample an API invocation from the Code LLM. This is used to retrieve documentation for the matching APIs. We then augment the prompt with the documentation and re-trigger the model.}
    \label{fig:dag-overview}
\end{figure}

\noindent \textbf{Overview.} Following~\citet{zhang2023repocoder, jiang2023active}, we implement an iterative pipeline for DAG. Starting with a prompt, the Code LLM generates a hypothetical API invocation. This invocation forms a query to retrieve documentation for similar APIs. The retrieved documentation is processed and appended to the original prompt, after which inference is re-triggered. This process is illustrated in Figure~\ref{fig:dag-overview}. 

\noindent \textbf{Query Formulation \& Retrieval Index.} Given a \bench task, the Code LLM generates a candidate API invocation, which we process as the query. This query focuses solely on API-relevant keywords, excluding any distractor prompt content~\citep{jiang2023active, zhang2023repocoder, eghbali2024dehallucinator}. Our retrieval index includes all collected AWS and Azure APIs, identified using \textit{keys} prepared similarly as the queries.

\noindent \textbf{Retriever.} We develop a retriever with configurable precision to study the effect of retrieval accuracy on \bench. For an $x\%$ precision@$k$ setting, we return $k$ documents via BM25, ensuring that the target API's documentation is included $x$\% of the time. This approach allows us to examine the impact of varying retrieval precision ($x$). We chose BM25 for its simplicity~\citep{patil2023gorilla, cheng2023lift}, though our results are likely robust to different retrievers.

\noindent \textbf{Augmentation Design.} We prepend the retrieved documentation to the original prompt as a Python docstring after postprocessing. We test various augmentation strategies, each capturing different levels of API information and token count efficiencies: (1) API Name Only, (2) API Description, (3) API Specification, (4) API Description + Specification, and (5) Full Documentation. Figure~\ref{fig:dag-api-specification} shows ``API Specification'' while additional details and examples are in Appendix~\ref{sec:appendix-augmentation-designs}.

\begin{figure}[t]
    \centering
    \includegraphics[width=0.85\columnwidth]{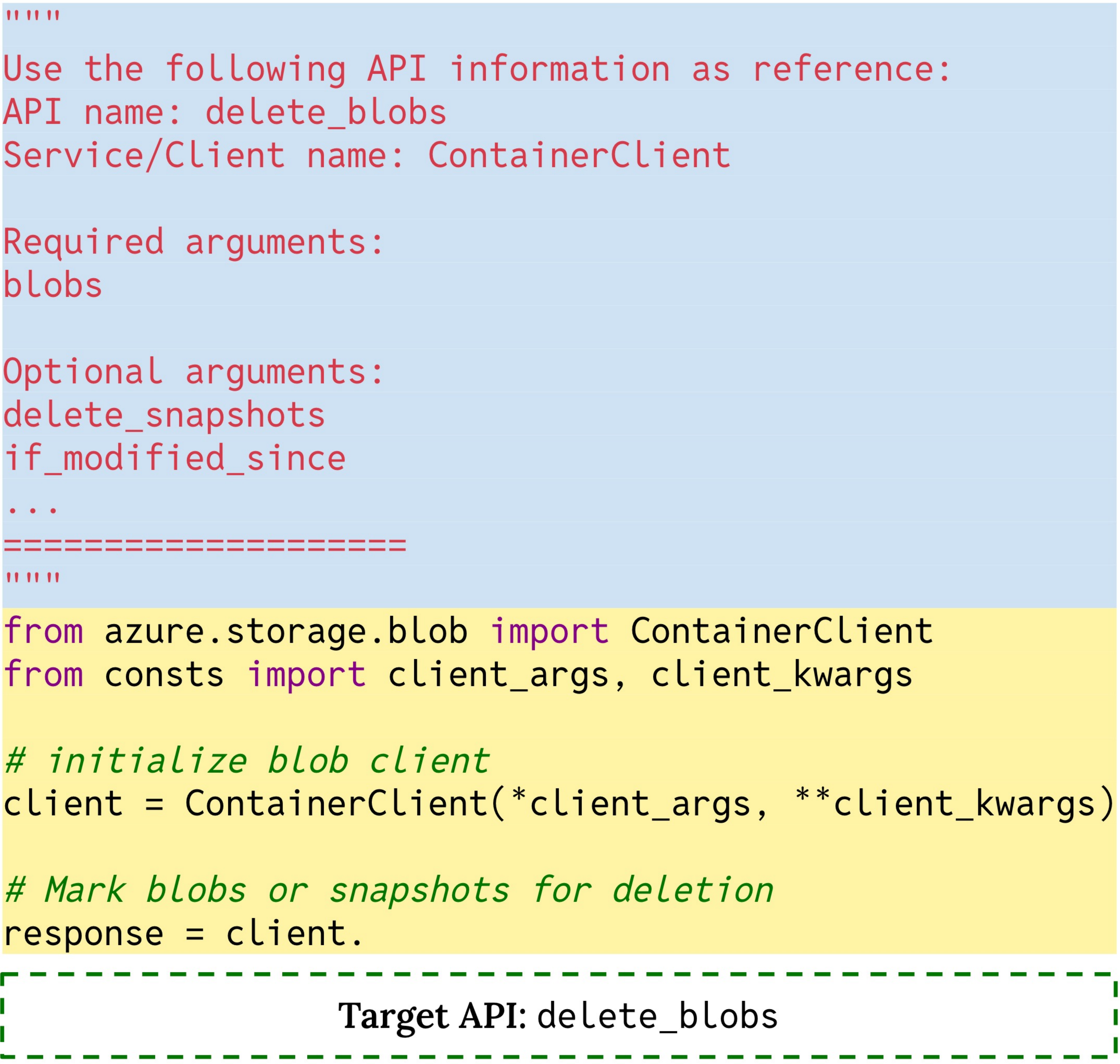}
    \caption{\textbf{API Specification Augmentation.} Augmented prompt for the Oracle retriever with one retrieval. The ``API Specification'' (blue) contains the API name and a list of its required \& optional arguments, providing an efficient summary of the documentation.}
    \label{fig:dag-api-specification}
\end{figure}

\subsection{Experiments \& Results} \label{sec:dag-experiments}

In this section, we perform ablations on various DAG components to analyze their impact on API hallucinations.

\noindent \textbf{Experimental Setup.} We present results from ablations on StarCoder2-3B. When testing a component (\eg, retriever precision), other components (\eg, number of retrievals) are held constant to isolate the effect. We also report the average valid API invocations across all tasks in \bench for a concise performance measure, wherever indicated.

\begin{figure*}[ht]
  \centering
  \begin{minipage}[]{0.48\textwidth}
    \centering
    \begin{table}[H]
\resizebox{\columnwidth}{!}{%
\renewcommand{\arraystretch}{1.2}
    \begin{tabular}{c|c|c}
    \hline
    \hline
    \textbf{Augmentation Design} & \textbf{Avg. Tokens} & \textbf{Valid API Inv. (\%)} \\
    \hline
    \hline
    \rowcolor{verylightgray} Base Model                                         & $--$ & $41.80$ \\
    \hdashline
    API Name Only                                           & $36.07$ & $52.73$ \\
    API Description                                           & $78.80$ & $53.22$ \\
    API Specification                                           & $52.57$ & \underline{$86.82$} \\
    API Desc. + API Spec.                                & $94.55$ & \underline{$87.14$} \\
    Full Documentation                                 & {\cellcolor{lightred}$685.24$} & $\mathbf{88.75}$ \\

    \hline
    \hline
    \end{tabular}
}
\end{table}

  \end{minipage}
  \hfill
  \begin{minipage}[]{0.48\textwidth}
    \centering
    \includegraphics[width=\textwidth]{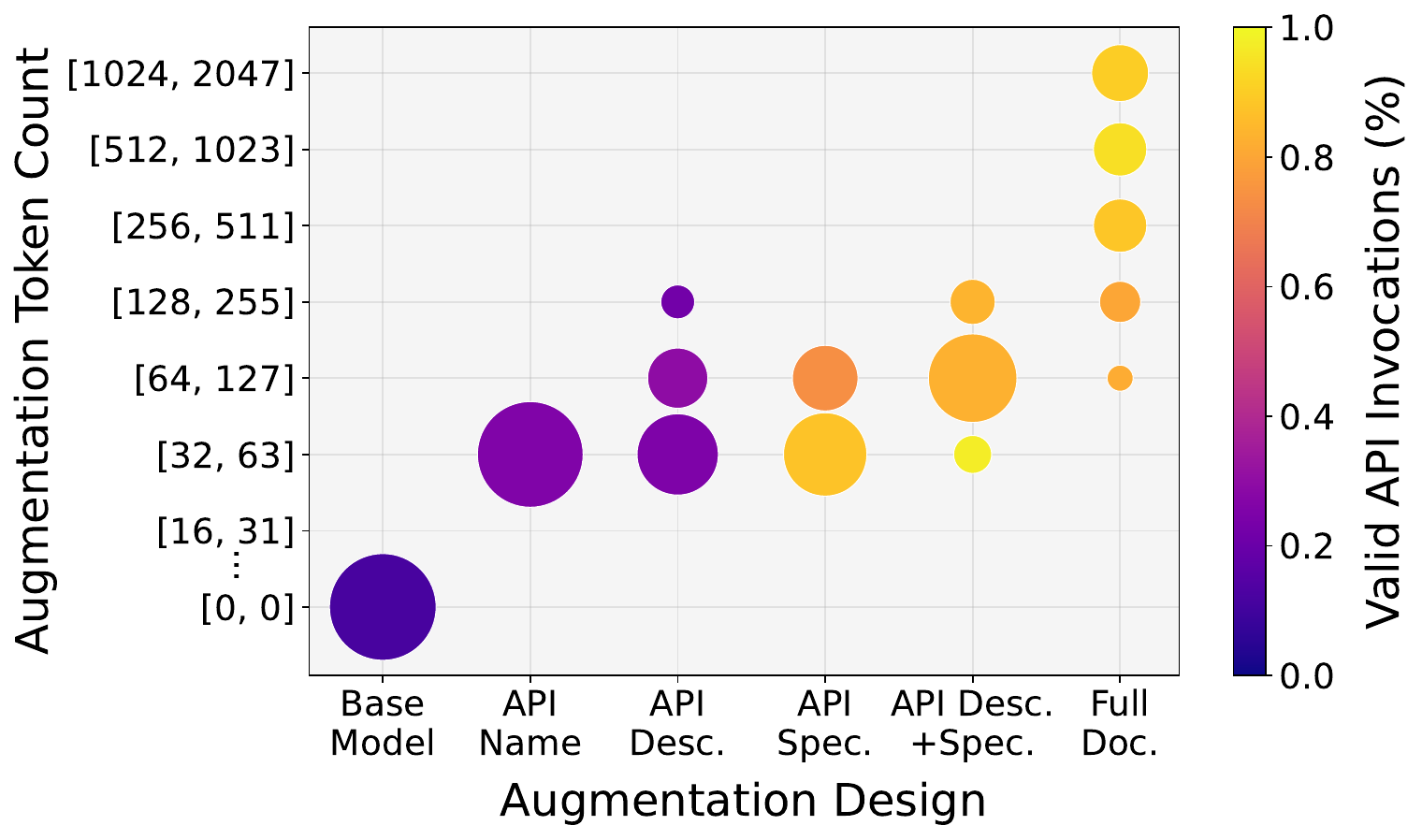} 
  \end{minipage}
  \caption{\textbf{Augmentation Design Results.} \textbf{(Left)}~Displays average tokens introduced per augmentation using the StarCoder2-3B tokenizer and average performance on \bench for each augmentation design. \textbf{(Right)}~Visualizes performance for \textit{low frequency} APIs: the \texttt{y-axis} shows binned sequence lengths (exponential scale; capped at $2048$), bubble color indicates performance, and bubble size indicates fraction of samples per bin. The improvements from API Specification are dramatic, though ``Full Documentation'' introduces too many tokens.}
  \label{fig:augmentation-design-results}
\end{figure*}

\noindent \textbf{Augmentation Design.} Our objective is to determine the most useful and efficient information to retain from an API's documentation for augmentation. So, we use an Oracle retriever to fetch only one documentation; this guarantees that the relevant information is present \textit{somewhere} in the documentation. Results are presented in Figure~\ref{fig:augmentation-design-results}, showing valid API invocation rates and the number of tokens introduced per augmentation across all APIs. ``API Name Only'' and ``API Description'' do not significantly reduce hallucination rates, as they lack detailed API syntax. However, adding ``API Specification'' dramatically improves model performance ($41.80\% \rightarrow 86.82\%$ on average), indicating that detailed API specifications are crucial. While ``Full Documentation'' optimizes performance, it is highly token-inefficient ($685.24$ tokens per augmentation). ``API Description + Specification'' strikes an optimal balance between token efficiency and performance, so, we adopt this design for all subsequent experiments.

\begin{figure}
    \centering
    \includegraphics[scale=0.225]{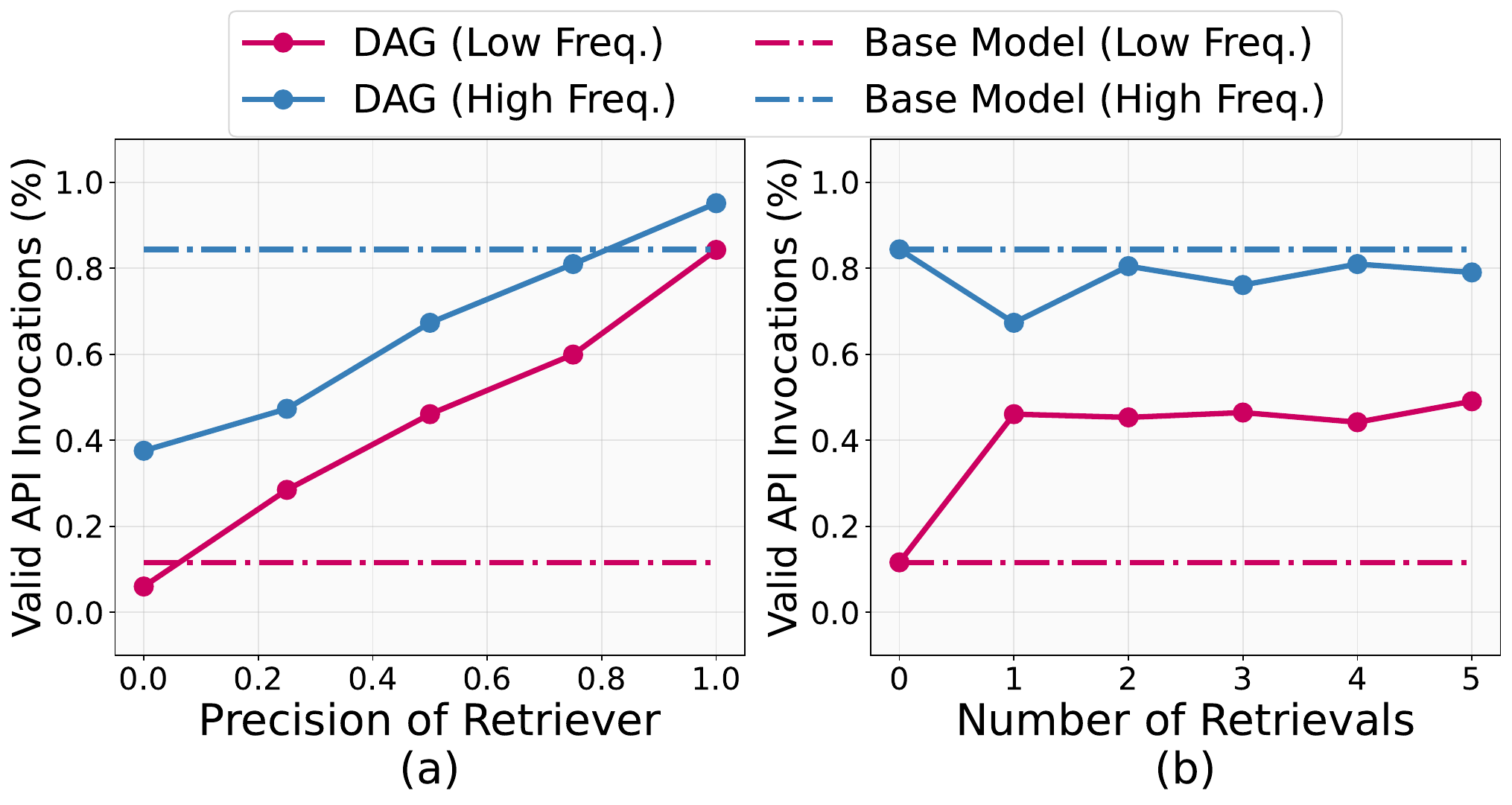}
    \caption{\textbf{Precision and No. of Retrievals.} (a)~While most low precision retrievers hurt performance on high frequency APIs, they may benefit low frequency APIs. (b)~$1$ retrieval hurts performance on high frequency APIs, but this is somewhat recovered as number of retrievals increases.}
    \label{fig:retrieval-results}
    \vspace{-5mm}
\end{figure}

\noindent \textbf{Precision of Retriever.} Results are displayed in Figure~\ref{fig:retrieval-results}a. Here, we retrieve one document and vary the retriever's precision. As anticipated, the API hallucination rate decreases as retriever precision increases. Low frequency APIs show improvement over the base model even with low precision retrievers, while high frequency APIs require precision $>80\%$ to match the base model's performance. Thus, at most precision levels, DAG induces higher hallucination rates for high frequency APIs compared to the base model (e.g., $84.39\% \rightarrow 67.32\%$ valid API invocations at $50\%$ precision), underscoring Code LLMs' sensitivity to irrelevant augmentations~\citep{shi2023large}.

\noindent \textbf{Number of Retrievals.} Here we maintain the retriever precision at $50\%$. Figure~\ref{fig:retrieval-results}b illustrates our findings. For low-frequency APIs, one or more retrievals consistently enhance performance. Conversely, high-frequency APIs show a sharp decline with one retrieval, partially recovered with two or more. This indicates that irrelevant augmentations can lead to unexpected behavior in Code LLMs, \textit{especially} when a single augmentation conflicts with the model's internal knowledge.

\noindent \textbf{Discussion.} Our experiments above show that DAG \textit{significantly} reduces hallucinations for low frequency APIs. However, high frequency APIs may suffer performance drops with DAG due to irrelevant retrievals. This issue can potentially be resolved by allowing the Code LLM to use its internal knowledge for high frequency APIs, bypassing DAG; this forms the core of the next section.

\section{Improving DAG: When to Retrieve?} \label{sec:improving-dag}

Given that suboptimal retrievers can increase hallucination rates with DAG, we investigate strategies to address this issue. By triggering DAG \textit{selectively} -- primarily when the Code LLM lacks knowledge of the target API -- we can mitigate the negative impact of suboptimal retrievals, and allow the model to invoke APIs correctly using its internal knowledge. We discuss two strategies towards this here.

\subsection{Index Lookup}

\noindent \textbf{Method.} This simple technique verifies if the API name invoked during the first iteration generation exists in the API index. If not, the Code LLM is trying to call a non-existing API, and DAG provides the necessary references. Thus, DAG is not triggered for existing APIs; since this is likely to happen for high-frequency APIs, we expect fewer imprecise DAG triggers with this method.

\noindent \textbf{Experimental Setup.} As before, we perform ablations with StarCoder2-3B. We use a $50\%$ precision retriever to retrieve one documentation.

\noindent \textbf{Results.} Table~\ref{tab:index-lookup-results} presents the results. The index lookup method significantly reduces the regressions introduced by DAG for high-frequency APIs, even showing slight improvements over the base model. However, this gain comes at the expense of reduced retrievals for low-frequency APIs: sometimes, the model invokes an existing incorrect API or incorrectly invokes the target API, leading to more hallucinations compared to DAG. Overall, this method shows promise for enhancing DAG.

\begin{table}
\centering
\resizebox{\columnwidth}{!}{%
\renewcommand{\arraystretch}{1.2}
    \begin{tabular}{c|ccc}
    \hline
    \hline
    \textbf{Method} & \textit{High Freq.} & \textit{Low Freq.} & \textit{Avg.} \\
    \hline
    \hline
    \rowcolor{verylightgray}Base Model                       &  $84.39$ & $11.61$ & $41.80$ \\
    \hdashline
    DAG                                                &  $67.32$ & $\mathbf{46.07}$ & $54.50$ \\
    DAG + Index Lookup                                 &  $\mathbf{85.37}$ & $35.96$ & $\mathbf{54.98}$ \\
    \hline
    \hline
    
    \end{tabular}
}%
\caption{\textbf{Index Lookup.} Triggering DAG only for non-existent APIs reduces unnecessary retrievals for high-frequency APIs, enhancing performance. However, this also induces slight regressions for low-frequency APIs. [\textit{Avg.} indicates performance across all frequencies.]}
\label{tab:index-lookup-results}
\end{table}

\subsection{API Invocation Confidence}

\textbf{Background: LLM Calibration.} Prior work has highlighted that LLM probabilities are well-calibrated, allowing for uncertainty estimation for various tasks~\citep{kadavath2022language, si2023prompting, jiang2023active, li2023web}. As such, leveraging a Code LLM's probabilities to predict potential hallucinations, we could selectively trigger DAG for those scenarios.

\begin{figure}
    \centering
    \includegraphics[width=\columnwidth]{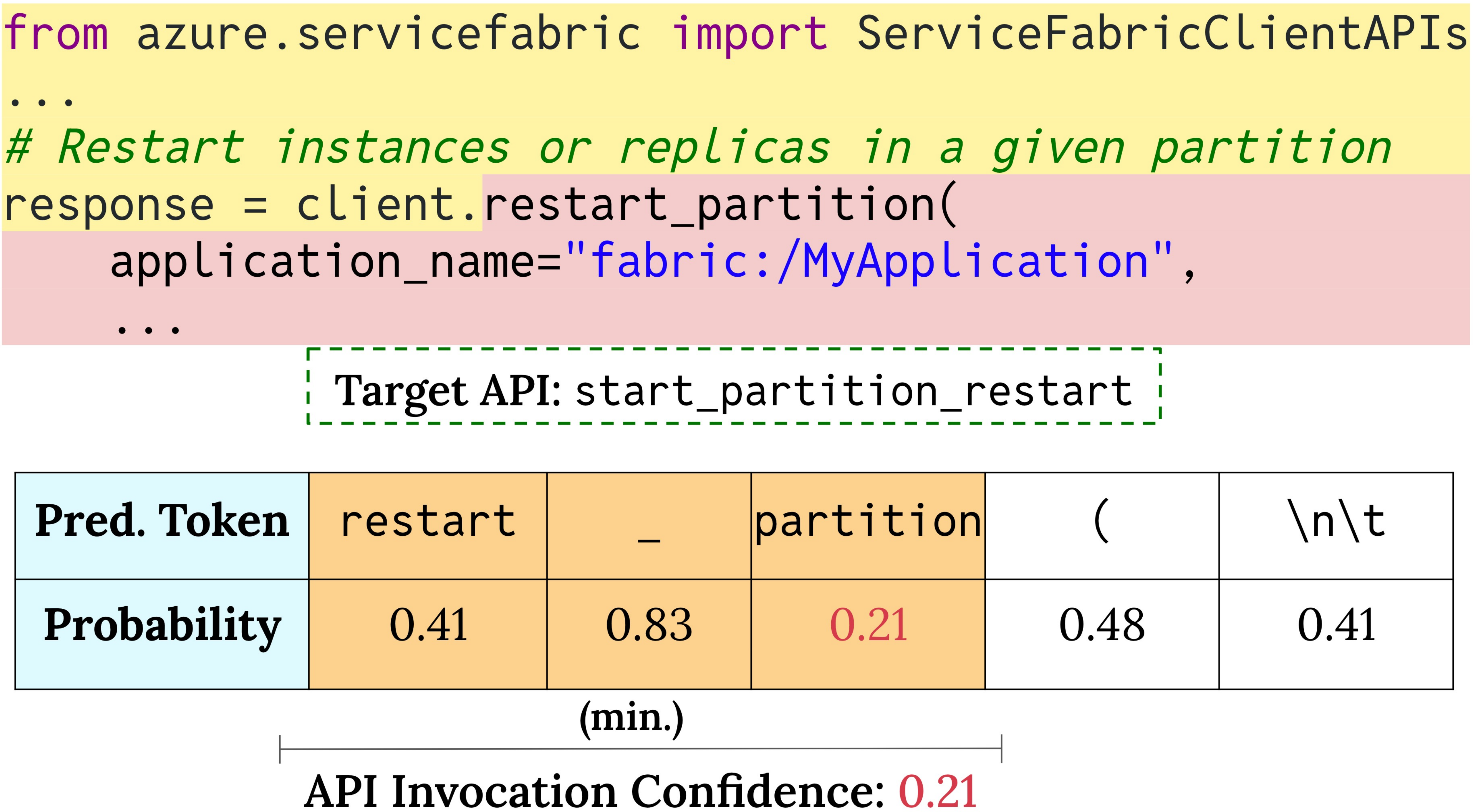}
    \caption{\textbf{API Invocation Confidence.} We estimate the model's uncertainty by taking the minimum probability of the predicted API name tokens (orange in table).}
    \label{fig:api-invocation-confidence}
\end{figure}

To quantify a Code LLM's uncertainty during API invocation, we define \textbf{API invocation confidence} as the \textit{minimum} probability among all predicted API name (sub-)tokens (see Figure~\ref{fig:api-invocation-confidence}). This minimum captures minor uncertainties in API prediction better than other aggregators like the mean~\citep{varshney2023stitch, jiang2023active}. The focus remains on the API name, not the entire invocation, as Code LLMs may show low confidence in tokens in the face of multiple alternatives (\eg, constants in API arguments, \etc; this represents aleatoric uncertainty~\cite{yadkori2024believe}).

\begin{figure}[ht]
    \centering
    \includegraphics[scale=0.225]{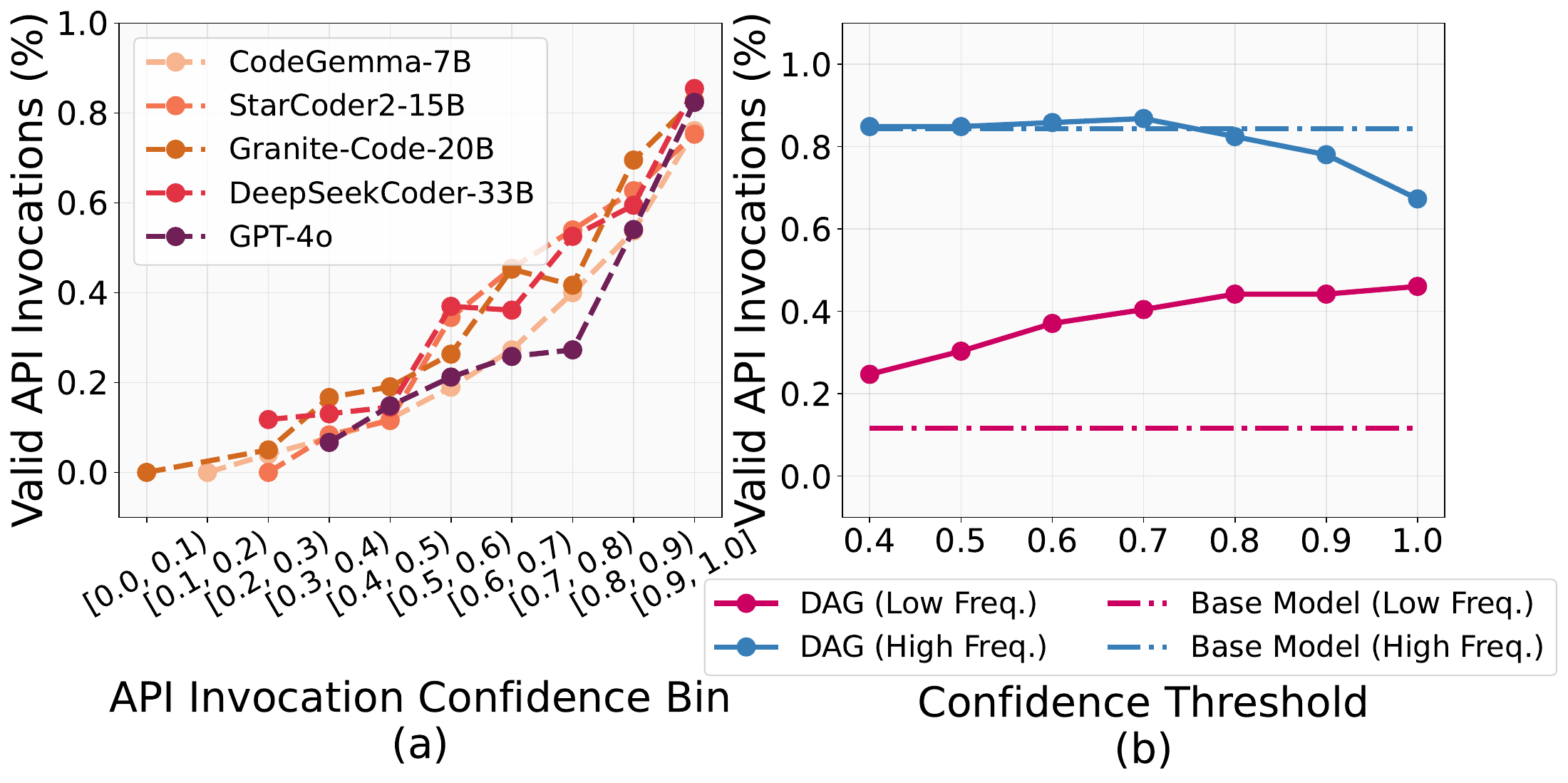}
    \caption{\textbf{API Invocation Confidence Results.} (a)~API invocation confidence scores are well-calibrated on \bench for a range of Code LLMs. (b)~By triggering DAG when the API invocation confidence is below a certain threshold, we can control regressions on high frequency APIs while maintaining good performance on low frequency APIs.}
    \label{fig:confidence-results}
    \vspace{-5mm}
\end{figure}

Evidence from various Code LLMs, shown in Figure~\ref{fig:confidence-results}a, confirms their calibration for API invocation on \bench. A strong positive correlation is observed between API invocation confidence and correct API usage, indicating that confidence levels can preemptively identify likely hallucinations (\ie, they capture epistemic uncertainty~\cite{yadkori2024believe}).

\noindent \textbf{Method.} We measure the API invocation confidence of the first iteration of generation, and if this is below a certain \textit{fixed threshold}, indicating the model's lack of knowledge about the target API, we trigger DAG to assist the model.

\noindent \textbf{Experimental Setup.} Towards finding an optimal configuration, we vary the threshold of API invocation confidence below which to trigger DAG, and measure the API hallucination rate for StarCoder2-3B. As before, we use a $50\%$ precision retriever with one retrieved document.

\noindent \textbf{Results.} Figure~\ref{fig:confidence-results}b shows the relation between the confidence threshold and valid API invocations. As we raise the threshold, DAG is triggered more often, leading to a consistent reduction in hallucinations for low frequency APIs. Conversely, high frequency APIs remain largely unaffected until a certain point, beyond which irrelevant augmentations start causing hallucinations. The optimal threshold balances improved performance for low frequency APIs without significant regressions for high frequency APIs; for StarCoder2-3B, this optimal range is approximately $0.7-0.8$.

\subsection{DAG++ \& Discussion}

\begin{table*}[ht]
\centering
\resizebox{\textwidth}{!}{%
\renewcommand{\arraystretch}{1.2}
    \begin{tabular}{c|c|ccc|cccc}
    \hline
    \hline
    \multirow{2}{*}{\textbf{Model}} & \multirow{2}{*}{\textbf{Method}} & \multicolumn{3}{c|}{\textbf{Retrieval Triggered (\%)}}  & \multicolumn{4}{c}{\textbf{Valid API Invocations (\%)}}\\
    & & \textit{High Freq.} & \textit{Med. Freq.} & \textit{Low Freq.} & \textit{High Freq.} & \textit{Med. Freq.} & \textit{Low Freq.} & \multicolumn{1}{:c}{\textit{Avg.}} \\
    \hline

\multirow{3}{*}{Google CodeGemma-7B} & \cellcolor{verylightgray}Base Model                                         &  \cellcolor{verylightgray}$0.00$ & \cellcolor{verylightgray}$0.00$ & \cellcolor{verylightgray}$0.00$ & \cellcolor{verylightgray}$87.80$ & \cellcolor{verylightgray}$52.67$ & \cellcolor{verylightgray}$12.36$ & \multicolumn{1}{:c}{\cellcolor{verylightgray}$46.95$} \\
& DAG                                                &  $100.00$ & $100.00$ & $100.00$ & $61.95_{(-25.85)}$ & $56.00_{(+3.33)}$ & $46.44_{(+34.08)}$ & \multicolumn{1}{:c}{$53.86_{(+6.91)}$} \\
& DAG++                                              &  $20.98$ & $44.67$ & $74.16$ & $88.29_{(+0.49)}$ & $65.33_{(+12.67)}$ & $43.07_{(+30.71)}$ & \multicolumn{1}{:c}{$\mathbf{63.34}_{(+16.40)}$} \\

\hdashline

\multirow{3}{*}{StarCoder2-15B} & \cellcolor{verylightgray}Base Model                                         &  \cellcolor{verylightgray}$0.00$ & \cellcolor{verylightgray}$0.00$ & \cellcolor{verylightgray}$0.00$ & \cellcolor{verylightgray}$88.78$ & \cellcolor{verylightgray}$57.33$ & \cellcolor{verylightgray}$24.72$ & \multicolumn{1}{:c}{\cellcolor{verylightgray}$53.70$} \\
& DAG                                                &  $100.00$ & $100.00$ & $100.00$ & $69.76_{(-19.02)}$ & $58.67_{(+1.33)}$ & $49.44_{(+24.72)}$ & \multicolumn{1}{:c}{$58.36_{(+4.66)}$} \\
& DAG++                                              &  $20.98$ & $43.33$ & $70.41$ & $88.78_{(+0.00)}$ & $58.67_{(+1.33)}$ & $46.44_{(+21.72)}$ & \multicolumn{1}{:c}{$\mathbf{63.34}_{(+9.65)}$} \\

\hdashline

\multirow{3}{*}{IBM Granite-Code-20B} & \cellcolor{verylightgray}Base Model                                         &  \cellcolor{verylightgray}$0.00$ & \cellcolor{verylightgray}$0.00$ & \cellcolor{verylightgray}$0.00$ & \cellcolor{verylightgray}$87.80$ & \cellcolor{verylightgray}$69.33$ & \cellcolor{verylightgray}$32.21$ & \multicolumn{1}{:c}{\cellcolor{verylightgray}$59.49$} \\
& DAG                                                &  $100.00$ & $100.00$ & $100.00$ & $70.24_{(-17.56)}$ & $63.33_{(-6.00)}$ & $44.19_{(+11.99)}$ & \multicolumn{1}{:c}{$57.40_{(-2.09)}$} \\
& DAG++                                              &  $15.12$ & $29.33$ & $66.29$ & $89.76_{(+1.95)}$ & $71.33_{(+2.00)}$ & $45.69_{(+13.48)}$ & \multicolumn{1}{:c}{$\mathbf{66.40}_{(+6.91)}$} \\

\hdashline

\multirow{3}{*}{DeepSeekCoder-33B} & \cellcolor{verylightgray}Base Model                                         &  \cellcolor{verylightgray}$0.00$ & \cellcolor{verylightgray}$0.00$ & \cellcolor{verylightgray}$0.00$ & \cellcolor{verylightgray}$90.24$ & \cellcolor{verylightgray}$70.00$ & \cellcolor{verylightgray}$34.83$ & \multicolumn{1}{:c}{\cellcolor{verylightgray}$61.58$} \\
& DAG                                                &  $100.00$ & $100.00$ & $100.00$ & $69.27_{(-20.98)}$ & $64.00_{(-6.00)}$ & $51.31_{(+16.48)}$ & \multicolumn{1}{:c}{$60.29_{(-1.29)}$} \\
& DAG++                                              &  $20.49$ & $30.67$ & $59.55$ & $86.83_{(-3.41)}$ & $71.33_{(+1.33)}$ & $55.43_{(+20.60)}$ & \multicolumn{1}{:c}{$\mathbf{69.61}_{(+8.04)}$} \\

\hdashline

\multirow{3}{*}{GPT-4o} & \cellcolor{verylightgray}Base Model                                         &  \cellcolor{verylightgray}$0.00$ & \cellcolor{verylightgray}$0.00$ & \cellcolor{verylightgray}$0.00$ & \cellcolor{verylightgray}$93.66$ & \cellcolor{verylightgray}$78.67$ & \cellcolor{verylightgray}$38.58$ & \multicolumn{1}{:c}{\cellcolor{verylightgray}$66.40$} \\
& DAG                                                &  $100.00$ & $100.00$ & $100.00$ & $54.63_{(-39.02)}$ & $53.33_{(-25.33)}$ & $47.94_{(+9.36)}$ & \multicolumn{1}{:c}{$51.45_{(-14.95)}$} \\
& DAG++                                              &  $3.41$ & $9.33$ & $50.56$ & $94.15_{(+0.49)}$ & $82.00_{(+3.33)}$ & $55.43_{(+16.85)}$ & \multicolumn{1}{:c}{$\mathbf{74.60}_{(+8.20)}$} \\

\hline
\hline

    \end{tabular}
}%
\caption{\textbf{DAG++ Results.} We present the performance on \bench and the (\%) of retrieval triggers for high/low frequency APIs, with absolute improvements over the base model shown in subscript. It is noteworthy that DAG++ significantly reduces the frequency of retrievals for high frequency APIs while appropirately decides to retrieve for low frequency APIs; by smartly triggering retrieval, DAG++ attains top performance on \bench for all models.} 
\label{tab:combined-results}
\end{table*}

Having seen the benefits of the above approaches, we now discuss how DAG can be effectively improved by combining these, \ie, DAG++. In this method, we trigger DAG \textit{iff} the API in the first iteration of generation does not exist in the index \textit{OR} is being invoked with an API invocation confidence below a fixed threshold. We anticipate that this  would help combine the benefits of both the discussed approaches.

\noindent \textbf{Experimental Setup.} We use a $50\%$ precision retriever with one retrieval, consistent with previous experiments. Further, we fix the confidence threshold to be $0.8$. Finally, to investigate the generalizability of our findings, we evaluate the largest models of all model families from Table~\ref{tab:no-augmentation-results}.

\noindent \textbf{Results.} The results are shown in Table~\ref{tab:combined-results}. For each model, show how often retrieval is triggered and the resulting performance on \bench. We make the following key observations:

\noindent -- \textit{Trigger of Retrievals.} We first examine how often retrieval is triggered with each method. The base model never triggers retrieval, DAG always does, and DAG++ selectively retrieves documentation. DAG++ exhibits a \textit{strong negative correlation} between retrieval trigger frequency and API frequency:~it triggers retrieval more often for low frequency APIs and less for high frequency APIs, aligning with the principle of retrieval \textit{only when necessary}. For \eg, with GPT-4o, DAG++ retrieves only $3.41\%$ of the time for high frequency APIs indicating minimal need for documentation; conversely, retrieval is triggered $50.56\%$ of the time for low frequency APIs, supplementing the model's limited knowledge with relevant documentation.

\noindent -- \textit{DAG v/s DAG++.} Table~\ref{tab:combined-results} also shows the performances (and absolute improvements over the base model in subscript) of various models on \bench. As noted in Section~\ref{sec:documentation-augmented-generation}, while DAG significantly boosts low frequency API performance, it degrades high frequency API performance. For instance, GPT-4o experiences a $39.02\%$ drop in performance for high frequency APIs with DAG, highlighting the the model's sensitivity to irrelevant augmentations. DAG++ successfully mitigates this issue for high frequency APIs while maintaining or improving gains on low frequency APIs. Overall, DAG++ outperforms DAG indicating that selective retrieval of API documentation, that respects API frequency, aids performance on \bench. 

\noindent -- \textit{Generalizability.} All model families demonstrate similar enhancement trends with DAG++, despite architectural and training differences. This underscores the generalizability of the importance of selectively retrieving API documentation when Code LLMs lack API specific knowledge. Additionally, scaling trends with model sizes~\citep{kaplan2020scaling, wang-etal-2023-shall} are evident: average performance monotonically improves with model size in Table~\ref{tab:combined-results}. Finally, DAG++ reveals that larger models require fewer retrievals for optimal performance, suggesting that they are more efficient at memorizing API syntax, even for low frequency APIs.

\section{Related Work}

\noindent \textbf{Program Synthesis \& API Invocations.} Code LLMs are actively being used for automatic program synthesis~\citep{rozire2023code, guo2024deepseekcoder}. Relevant to our study is API invocation generation~\citep{qin2023toolllm, patil2023gorilla}, often done on tool-usage benchmarks that do not account for the distribution of APIs in the public domain. We develop \bench, a benchmark targeting cloud-based software engineering scenarios, that includes API frequency annotations, allowing for nuanced failure analyses and targeted improvements through DAG. Works such as~\citet{zhou2023docprompting, patil2023gorilla, eghbali2024dehallucinator, zan2023privatelibraryoriented} also use documentation to improve API generation, but their evaluations do not capture the granularities discussed here.

\noindent \textbf{LLM Hallucinations.} LLMs may generate factually incorrect statements about concepts, diminishing their utility~\citep{mishra2024finegrained, kang2023mitigating, lee2023large}. As such, several works have emerged to deal with this issue. Some works focus on hallucination detection by exploiting the well-calibrated nature of LLMs~\citep{kadavath2022language, si2023prompting, li2023web} and using model confidence scores~\citep{jiang2023active, varshney2023stitch}. Closest to our work, \citet{liu2024exploring}, give a taxonomy of hallucinations for code generation. While they focus on identifying hallucinations with Code LLMs, we focus on mitigating API hallucinations using documentation.

\noindent \textbf{Retrieval Augmented Generation (RAG).} RAG supplements language models by retrieving from external data-stores~\citep{asai-etal-2023-retrieval}. Some studies use fixed algorithms for retrieval~\citep{wang2023selfknowledge, shi2023replug, patil2023gorilla}, while others adopt adaptive retrieval through special tokens~\citep{asai2023selfrag} or model confidence scores~\citep{jiang2023active}. In this work, we establish how to use selective retrieval effectively to mitigate API hallucinations with documentation.

\section{Conclusion \& Future Work} \label{sec:conclusion}

In this work, we thoroughly investigate API hallucinations and demonstrate mitigation strategies for various Code LLMs. We introduce \bench, a benchmark to measure API hallucinations for diverse AWS and Azure APIs, including API frequencies to categorize low, medium, and high frequency APIs. We adapt RAG with documentation (DAG) to inform Code LLMs about the correct syntax during inference. We discuss which parts of documentation are important and how various retrieval components affect hallucinations. While DAG significantly enhances low-frequency API performance, it can degrade high-frequency API performance with irrelevant retrievals. We tackle this issue by \textit{selectively} triggering retrievals through index lookup and API invocation confidence thresholding, and combine these methods in DAG++ leading to top performance on \bench across Code LLMs. Future research could extend \bench for long-context evaluations, explore DAG beyond iterative generation, and improve DAG by enhancing Code LLMs' robustness to irrelevant augmentations.

\section{Limitations} \label{sec:limitations}

\noindent \textbf{Scope of \bench.} \bench is a Python only benchmark containing short synthetic prompts to evaluate API hallucinations. While these represent various software-engineering scenarios, these might not represent all real-world cloud API invocations across different programming languages and contexts.  

\noindent \textbf{Construction of \bench.} We create \bench using a multi-step process as discussed in Section~\ref{sec:cloudapibench} and Appendix~\ref{sec:appendix-cloudapibench-construction}. Some of these steps are based on carefully crafted heuristics such as a customized logic to estimate API frequencies. Given that our findings are consistent with literature and also match our expectations, the impact of the approximations employed, if any, should be limited.

\noindent \textbf{Iterative generations for DAG.} In this work, we have adopted an iterative approach to DAG where we generate, retrieve and re-generate. Due to the overhead introduced by this iterative process, it may not be suitable for scenarios where latency is crucial.

\section{Ethics Statement} \label{sec:ethics-statement}

\noindent \textbf{Use of Generative AI.} Code generation models are subject to ethical risks as these models can generate harmful content or content similar to their pre-training data. For real world applications, the generated content should ideally be reviewed by human developers and should be executed in sandbox environments. For the scope of experiments in this work, these risks are relatively low.

\noindent \textbf{Compute.} Use of deep learning models is computationally expensive and raises environmental concerns. We have not trained any models as part of this work, so, the computational footprint is relatively low. All experiments for this paper were done using $4$ NVIDIA A100 machines.



\bibliography{custom}

\twocolumn[{%
    {\centering
    \Large\bf Supplementary Material: Appendices \\ [20pt]}
}]
\appendix
\section{\bench} \label{sec:appendix-cloudapibench}

\subsection{Composition} \label{sec:appendix-cloudapibench-composition}

\begin{figure}
    \centering
    \includegraphics[width=0.95\columnwidth]{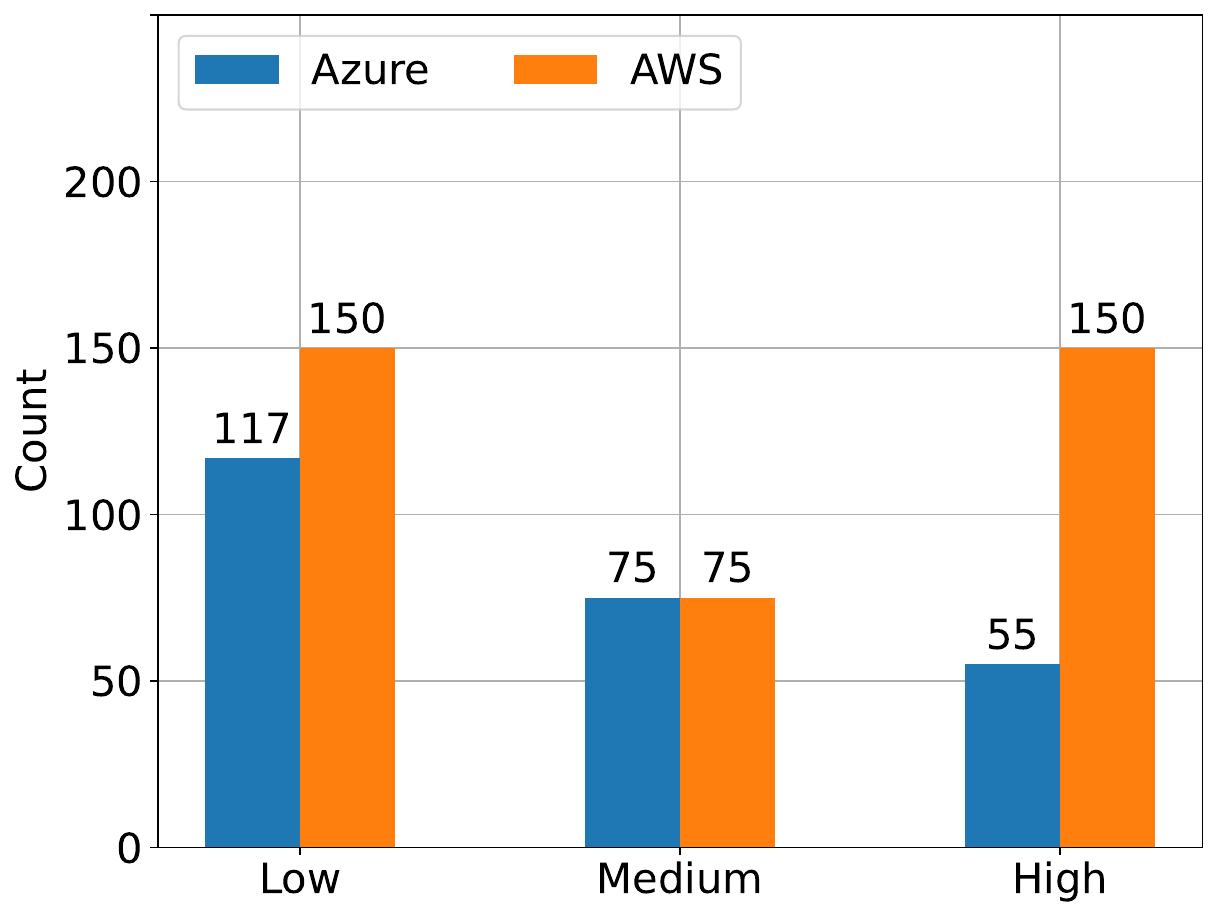}
    \caption{\textbf{\bench API Frequency Distribution.} \texttt{y-axis} shows number of tasks in \bench satisfying the respective criterion.}
    \label{fig:appendix-cloudapibench-dist}
\end{figure}

We give more details about the composition of \bench here. \bench comprises several AWS and Azure APIs, each annotated with an API frequency proportional to the API's representation in public sources. Figure~\ref{fig:appendix-cloudapibench-dist} shows the API frequency distribution across three bins: low, medium and high, for both AWS and Azure.

\subsection{Construction} \label{sec:appendix-cloudapibench-construction}

\begin{figure}[H]
    \centering
    \includegraphics[width=\columnwidth]{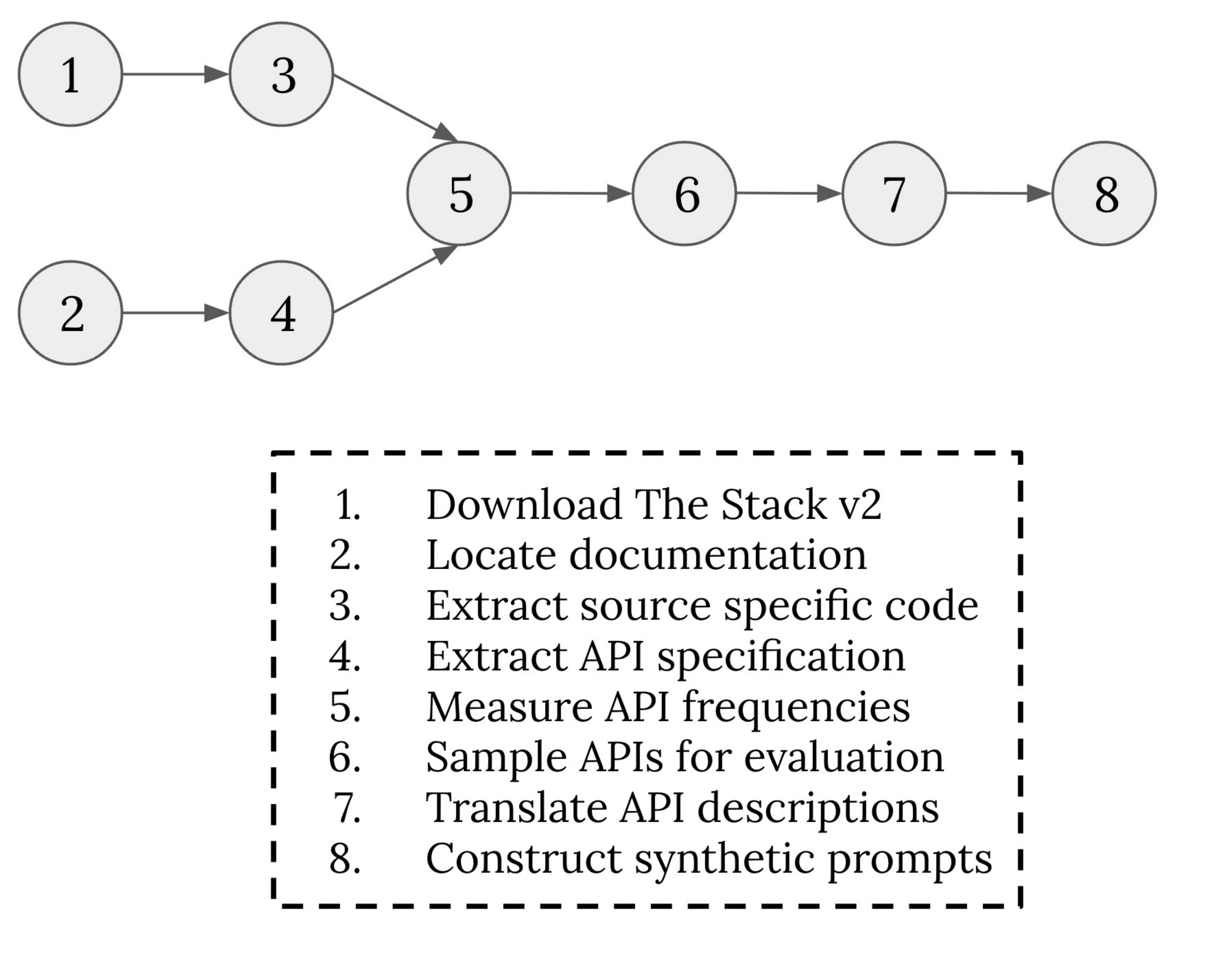}
    \caption{\textbf{Summary of steps to construct \bench.}}
    \label{fig:appendix-cloudapibench-construction}
\end{figure}

We follow a multi-step procedure to obtain synthetic evaluation tasks in Python for AWS and Azure APIs to include in \bench. A summary of the process is shown in Figure~\ref{fig:appendix-cloudapibench-construction}. We describe each step in detail here.

\begin{enumerate}
    \item \textbf{Download The Stack v2.} We download The Stack v2 from its official repository on HuggingFace and SoftwareHeritage~\citep{lozhkov2024starcoder}.
    
    \item \textbf{Locate Documentation \& Syntax.} We use \texttt{boto3 1.34.108} for AWS and the Python package \texttt{azure-sdk-for-python} for Azure. For AWS, we use the \texttt{mypy\_boto3\_builder} tool~\citep{mypyboto3-builder} to create API stubs for all AWS APIs; this helps us obtain the list of APIs. We obtain the official documentation for each of these by scraping the \texttt{boto3} webpage for the respective APIs. For Azure, the complete docstring in the source code for an API's defintion is its documentation.
    
    \item \textbf{Extract source specific code.} We identify source specific code samples in The Stack v2 so that we restrict the count of API frequencies to only these. For AWS, source specific files are those that import one of ~\texttt{\{boto3, botocore\}} or contain one of \texttt{\{aws, boto, amazon\}} in the filepath. Similarly, Azure specific samples are those that import \texttt{azure} or contain \texttt{azure} in the filepath.
    
    \item \textbf{Extract API specification.} For Azure, the complete documentation is available as a docstring in the respective function definitions for that API. Using tree-sitter, we parse the code files to obtain the list of APIs, their correct usages and complete docstrings for Azure, for as many APIs as possible. For AWS, we parse API stubs obtained using \texttt{mypy\_boto3\_builder} to curate the API specifications. This also serves as the index of APIs that we use in our experiments.
    
    \item \textbf{Measure API frequencies.} Given the list of APIs for a source, we count the number of times functions with the name as an API are invoked or defined within the source specific code samples identified above. We use several heuristics to avoid edge cases and maintain reliability. Nevertheless, some noise may creep in and we acknowledge that this process is far from perfect. However, the findings based off of these API frequencies align with our expectations, indicating their reliability.

    \item \textbf{Sample APIs for evaluation.} While sampling APIs for evaluation, we take care to ensure diversity with respect to API frequencies (uniform sampling from each frequency class as far as possible) and invocation complexity (within each frequency class, there should be uniform distribution of the number of required arguments required by APIs). This ensures that \bench is diverse and represents diverse software-engineering scenarios, from low to high frequency, and from APIs that are easy to invoke to those that require careful recall of API syntax. Further each API may appear in \bench up to $3$ times with different prompts.

    \item \textbf{Translate API descriptions.} Each sample in \bench contains a comment that expresses the intent to invoke the API. We obtain this comment by instructing Claude 3 Sonnet~\cite{Claude3} to translate the documentation description of the API into $3$ concise developer style comments. We use few-shot prompting to do this, and upon manual inspection of dozens of responses, find that Claude is able to do this task reliably. As such, we use Claude's responses as comments describing the intent to invoke the APIs, fixing any issues that were noticed manually.

    \item \textbf{Construct synthetic prompts.} As the final step, for the selected APIs, we create synthetic prompts by creating an incomplete code sample: these start with relevant imports, contain necessary variable declarations, include the comment expressing the intent to invoke an API, and end just before the API invocation. Manual inspection revealed that in a few cases, multiple APIs may be suitable targets for a task, and in such cases we manually enumerate all the possible targets to the best of our knowledge.
    
\end{enumerate}

\begin{table*}
\centering
\resizebox{\textwidth}{!}{%
\renewcommand{\arraystretch}{1.1}
    \begin{tabular}{c|cc|ccc}
    \hline
    \hline
    \multirow{2}{*}{\textbf{Model}} & \multirow{2}{*}{\textbf{Valid (\%)}} & \multirow{2}{*}{\textbf{Invalid (\%)}} & \multicolumn{3}{c}{\textbf{Invalid API Invocations Breakdown}} \\
    & & & \textit{Usage of incorrect existing} & \textit{Invalid usage of target} & \textit{Usage of non-existing} \\
    \hline
    Google CodeGemma-7B                                & $12.36$ & $87.64$ & $10.68$ & $35.47$ & $53.85$ \\
    StarCoder2-15B                                     & $24.72$ & $75.28$ & $10.95$ & $33.33$ & $55.72$ \\
    IBM Granite-Code-20B                               & $32.21$ & $67.79$ & $17.13$ & $23.76$ & $59.12$ \\
    DeepSeekCoder-33B                                  & $34.83$ & $65.17$ & $15.52$ & $31.03$ & $53.45$ \\
    GPT-4o                                             & $38.58$ & $61.42$ & $13.41$ & $33.54$ & $53.05$ \\
    \hline
    \hline
    \end{tabular}
}%
\caption{\textbf{Results on \bench for low frequency APIs.} We first show the fraction of valid and invalid API invocations for low frequency APIs for various models. The invalid API invocations are categorized into various types of failures. Notably, $> 50\%$ failures occur due to the models attempting to invoke non-existing APIs.}
\label{tab:appendix-low-freq-deep-dive}
\end{table*} 

\subsection{Evaluation} \label{sec:appendix-cloudapibench-evaluation}

\noindent \textbf{Metric Calculation.} When more than one target API is identified, as described in Appendix~\ref{sec:appendix-cloudapibench-construction}, we consider the candidate to be valid as long as it satisfies the syntax of any one of the target APIs. \\

\noindent \textbf{Inference.} While base models can be directly evaluated on \bench, instruction-tuned models need to be instructed to generate an API invocation. We use a system prompt to achieve this; this is shown in Listing~\ref{lst:appendix-evaluation-system-prompt}.

\lstset{style=systemprompt}
\begin{lstlisting}[language={}, float=*, label=lst:appendix-evaluation-system-prompt, caption=System prompt to evaluate instruction-tuned models such as GPT-4o on \bench.] 
You are code completion model. You generate code starting from the end of the prompt given to you. You will give your output surrounded by backticks.

Notably, the prompt requires you to complete an API invocation. Complete the API invocation and stop there. Do not write any code other than the single API invocation.

As an example you will be given a code input. And you should return your output as:
```python
<API_INVOCATION_HERE>
```
\end{lstlisting}

\subsection{\bench Samples} \label{sec:appendix-cloudapibench-samples}

\begin{figure*}
    \centering
    \includegraphics[width=0.9\textwidth]{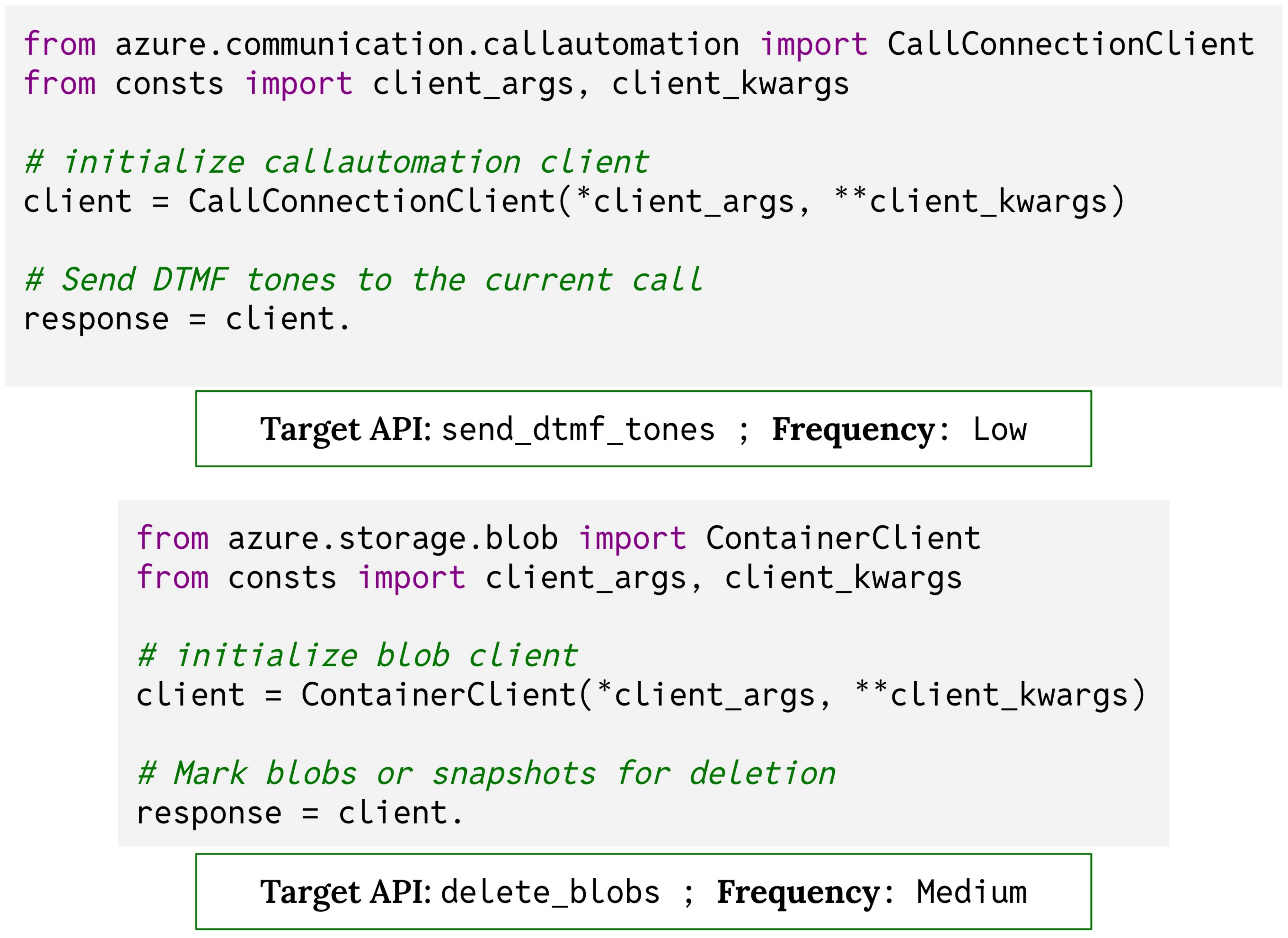}
    \caption{\textbf{Azure samples from \bench.}}
    \label{fig:appendix-azure-samples}
\end{figure*}

\begin{figure*}
    \centering
    \includegraphics[width=0.8\textwidth]{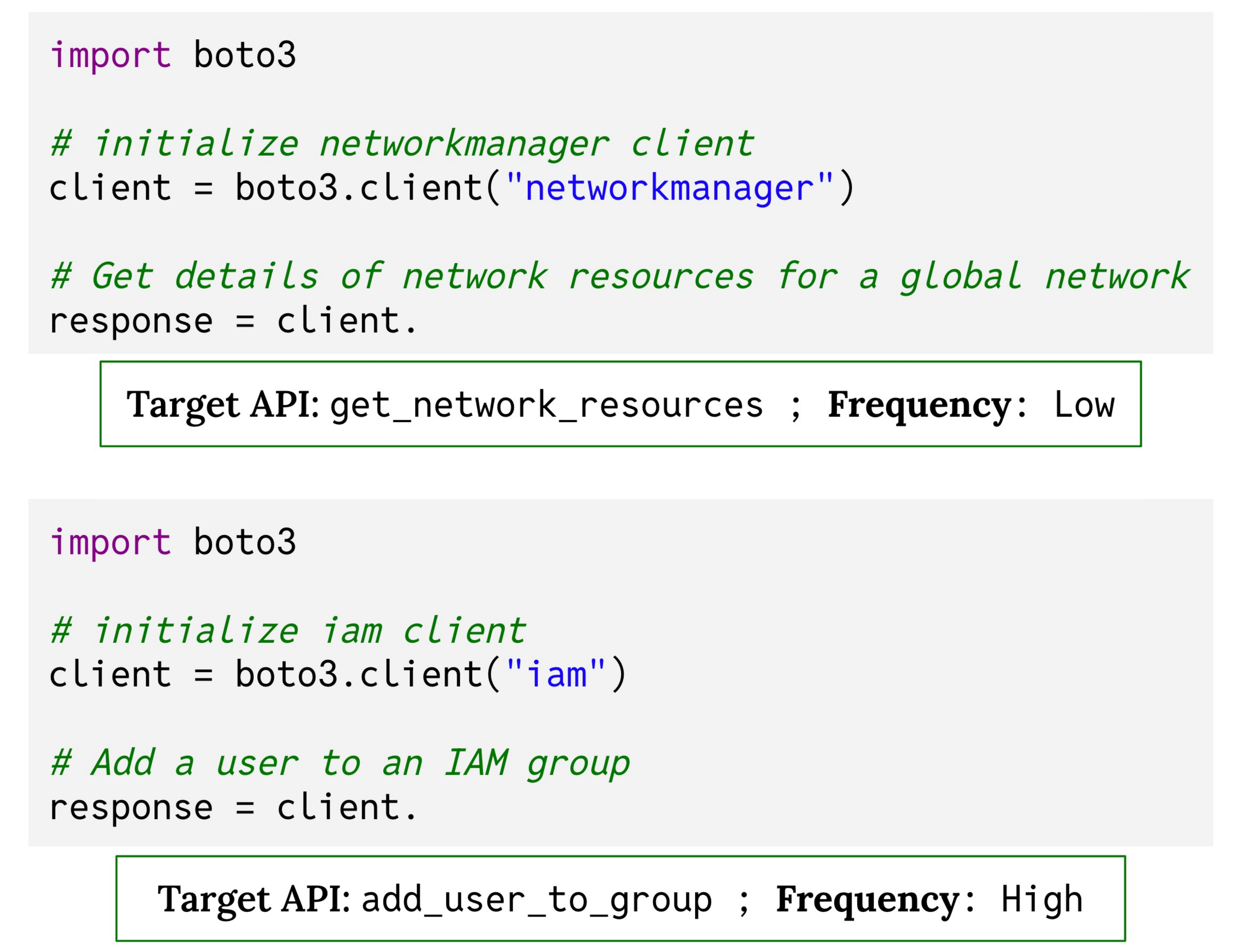}
    \caption{\textbf{AWS samples from \bench.}}
    \label{fig:appendix-aws-samples}
\end{figure*}

We show more samples from \bench for illustration purposes here. Azure samples are shown in Figure~\ref{fig:appendix-azure-samples} and AWS samples are shown in Figure~\ref{fig:appendix-aws-samples}. Each sample also shows the target API and the frequency classification of the API.

\subsection{Hallucination Categorization \& Illustration} \label{sec:appendix-cloudapibench-results}

\begin{figure*} 
    \centering
    \begin{subfigure}[b]{0.8\textwidth}
         \centering
         \includegraphics[width=\textwidth]{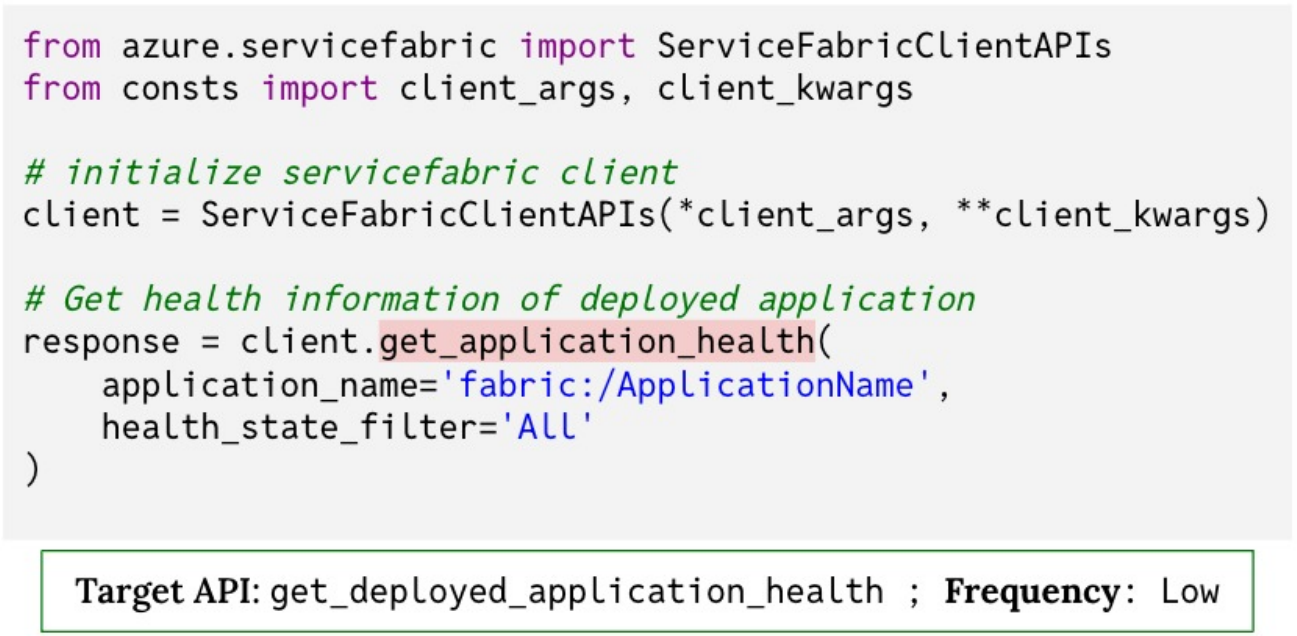}
         \caption{The model is attempting to invoke an API that exists (\texttt{get\_application\_health}) but does not match the task description for the target API (\texttt{get\_deployed\_application\_health}).}
         \label{fig:appendix-hallucinations-1}
     \end{subfigure}
     \begin{subfigure}[b]{0.8\textwidth}
         \centering
         \includegraphics[width=\textwidth]{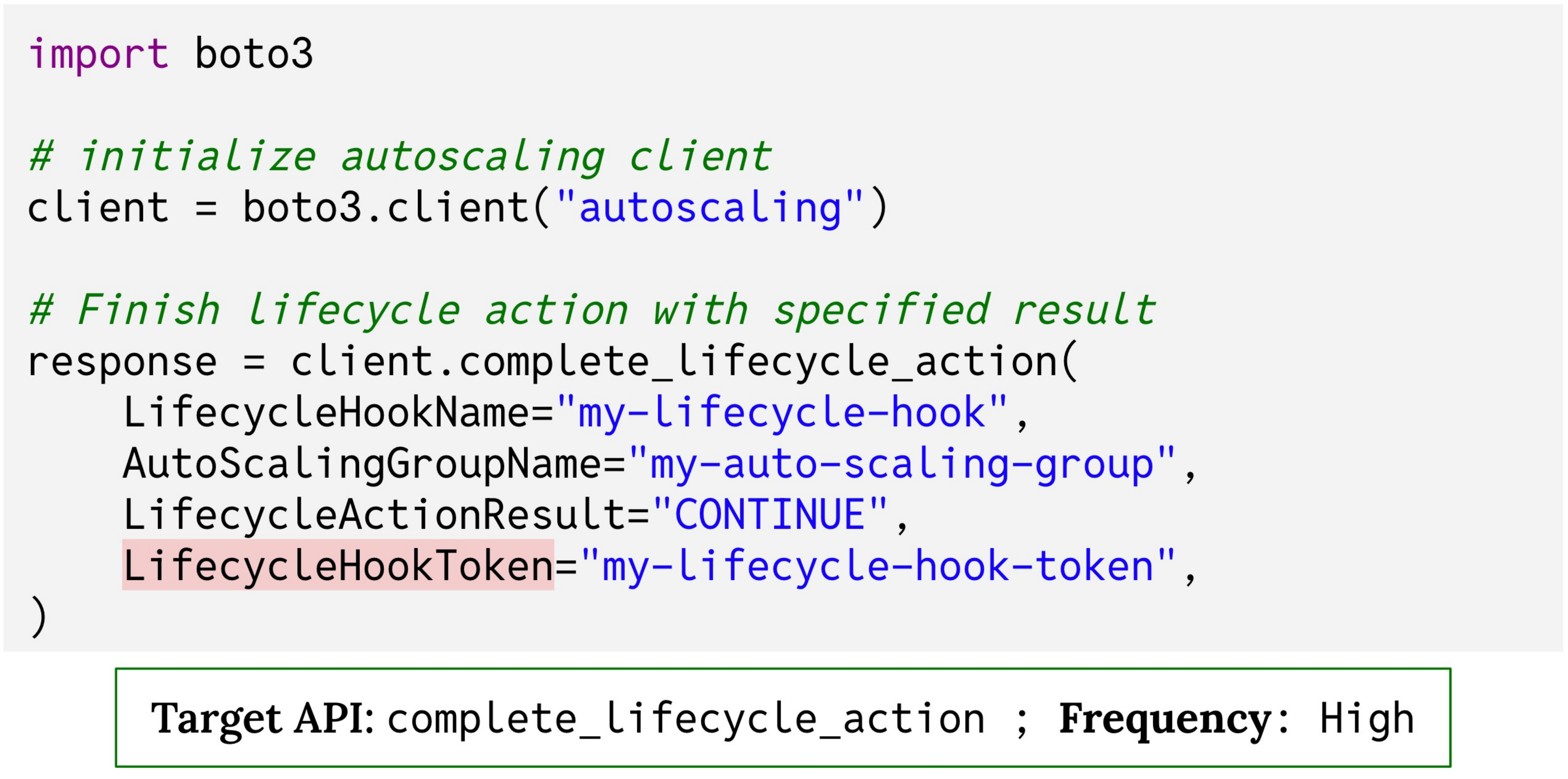}
         \caption{For this high frequency API, the model passes an argument that the API does not accept (red). Notably, all other arguments passed here are valid.}
         \label{fig:appendix-hallucinations-2}
     \end{subfigure}
     \begin{subfigure}[b]{0.8\textwidth}
         \centering
         \includegraphics[width=\textwidth]{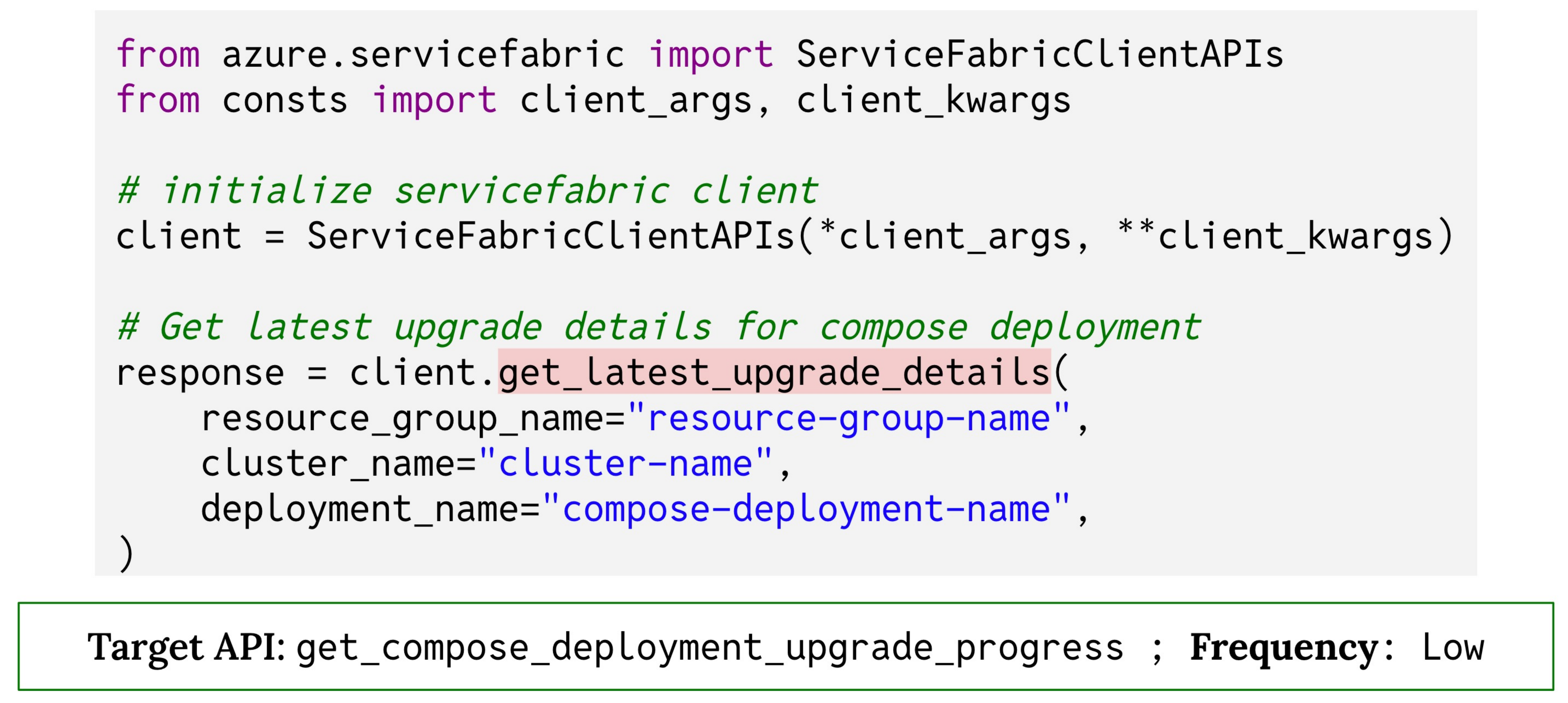}
         \caption{The model is attempting to invoke a non-existent API here.}
         \label{fig:appendix-hallucinations-3}
     \end{subfigure}
     \caption{\textbf{API Hallucination Scenarios.} We show three different ways in which Code LLMs hallucinate for tasks on \bench. Here the model responses are from Google CodeGemma-7B.}
     \label{fig:appendix-hallucinations}
\end{figure*}

We classify API hallucinations into three broad categories:

\begin{enumerate}
    \item \textbf{Usage of incorrect existing API}. This occurs when the Code LLM attempts to invoke an API that exists but does not fulfill the task (see Figure~\ref{fig:appendix-hallucinations-1}).
    \item \textbf{Invalid usage of target API}. This occurs when the Code LLM attempts to invoke the correct API but does so incorrectly due to an invalid configuration of arguments; here the model may either pass arguments that the API does not accept or not pass a correct combination of required and optional arguments (see Figure~\ref{fig:appendix-hallucinations-2}).
    \item \textbf{Usage of non-existing API}. This occurs when the Code LLM attempts to invoke an API that does not exist (see Figure~\ref{fig:appendix-hallucinations-3}).
\end{enumerate}

\subsection{Analyzing Low Frequency API Failures} \label{sec:appendix-cloudapibench-low-freq-deep-dive}

We look closer into the various modes of failure for low frequency APIs in Table~\ref{tab:appendix-low-freq-deep-dive}. We present this analysis for the largest model in each model family.

As shown, most failures arise when the models try to invoke a non-existing API or use the target API incorrectly. This goes to show the lack of knowledge about low frequency APIs, and the propensity to hallucinate under these scenarios in Code LLMs.

\pagebreak
\section{Documentation Augmented Generation (DAG)} \label{sec:appendix-dag}

\subsection{Augmentation Designs} \label{sec:appendix-augmentation-designs}

We define and illustrate various augmentation designs in this section.

\begin{itemize}
    \item \textbf{API Name Only.} We include only the name of the retrieved APIs as augmentation. This can test if the Code LLM can invoke the correct API just by referencing its name during inference.
    \item \textbf{API Description.} We include the name and a short description of the API. For Azure the short description is the first sentence from the API's docstring, whereas for AWS the short description is the first $5$ sentences from the API's documentation on the \texttt{boto3} webpage. We choose $5$ here as we found, in several cases, the first $2-3$ sentences to be irrelevant to the API's description.
    \item \textbf{API Specification.} This is a concise summary of the syntax of the API. It includes the name of the API and the list of required and optional arguments without specifying any descriptions of the arguments.
    \item \textbf{API Description + API Specification.} This includes the description as defined above along with the specification as discussed above.
    \item \textbf{Full Docstring.} This uses the entire collected documentation as augmentation. Since this can be arbitrarily large, especially for AWS documentation, we right-truncate the documentation up to $5000$ characters before augmenting. This assumes that the necessary information to invoke the API is within the first $5000$ characters.
\end{itemize}

\begin{figure*}
    \centering
    \includegraphics[width=1.\textwidth]{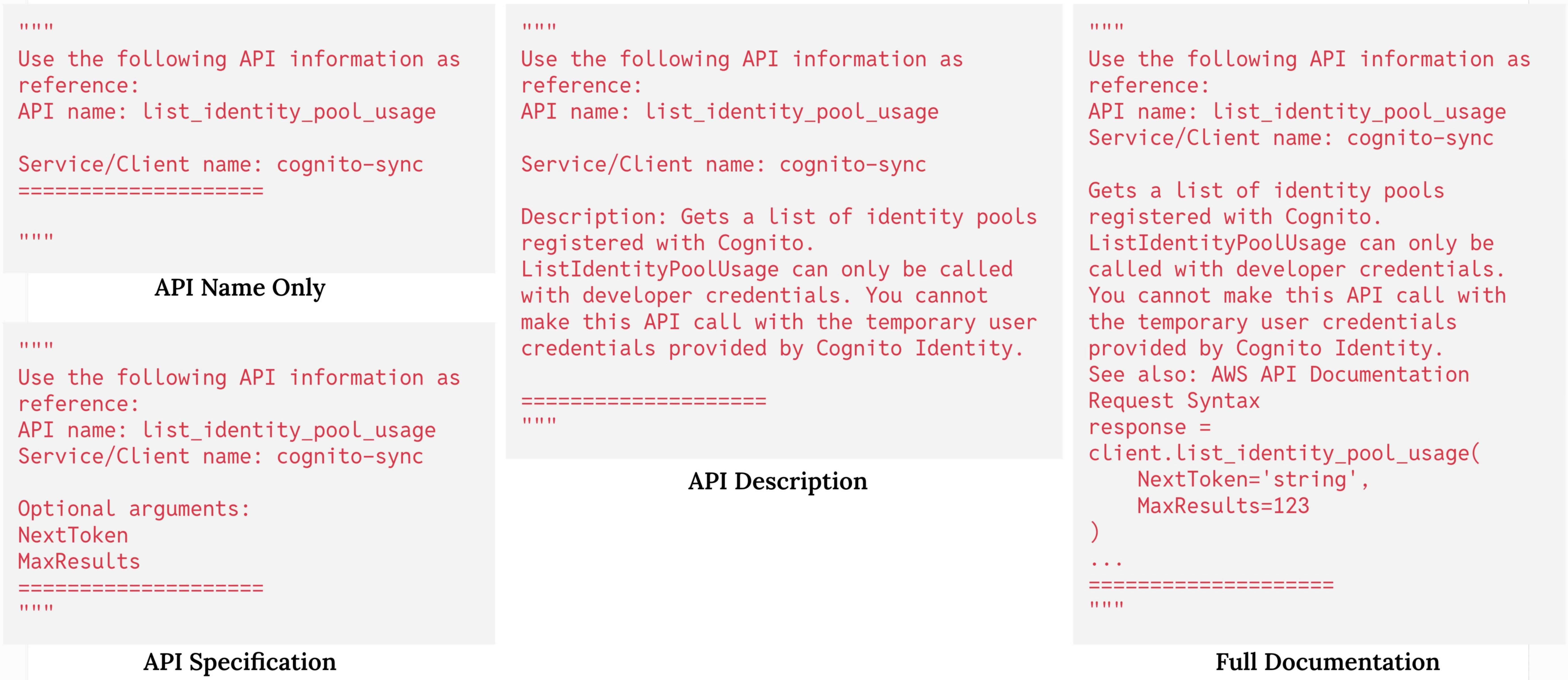}
    \caption{\textbf{API augmentation designs}. Illustrated for the AWS API: \texttt{list\_identity\_pool\_usage}. ``Full Documentation'' is truncated to fit in the figure.}
    \label{fig:appendix-aws-augmentation}
\end{figure*}

We illustrate these strategies in Figure~\ref{fig:appendix-aws-augmentation}. We skip ``API Description + API Specification'' in the figure as it is a combination of ``API Description'' and ``API Specification''.

\end{document}